\PassOptionsToPackage{unicode,hypertexnames=false}{hyperref}
\PassOptionsToPackage{hyphens}{url}
\PassOptionsToPackage{dvipsnames,svgnames,x11names}{xcolor}
\documentclass[
  12pt]{article}

\usepackage{amsmath,amssymb,amsthm}
\usepackage{iftex}
\ifPDFTeX
  \usepackage[T1]{fontenc}
  \usepackage[utf8]{inputenc}
  \usepackage{textcomp} 
\else 
  \usepackage{unicode-math}
  \defaultfontfeatures{Scale=MatchLowercase}
  \defaultfontfeatures[\rmfamily]{Ligatures=TeX,Scale=1}
\fi
\usepackage{lmodern}
\ifPDFTeX\else  
\fi

\IfFileExists{upquote.sty}{\usepackage{upquote}}{}
\IfFileExists{microtype.sty}{%
  \usepackage[protrusion=true,expansion=true]{microtype}
  \UseMicrotypeSet[protrusion]{basicmath}
}{}{}
\makeatletter
\@ifundefined{KOMAClassName}{
  \IfFileExists{parskip.sty}{%
    \usepackage{parskip}
  }{
    \setlength{\parindent}{0pt}
    \setlength{\parskip}{6pt plus 2pt minus 1pt}}
}{
  \KOMAoptions{parskip=half}}
\makeatother
\usepackage{xcolor}
\makeatletter
\ifx\paragraph\undefined\else
  \let\oldparagraph\paragraph
  \renewcommand{\paragraph}{
    \@ifstar
      \xxxParagraphStar
      \xxxParagraphNoStar
  }
  \newcommand{\xxxParagraphStar}[1]{\oldparagraph*{#1}\mbox{}}
  \newcommand{\xxxParagraphNoStar}[1]{\oldparagraph{#1}\mbox{}}
\fi
\ifx\subparagraph\undefined\else
  \let\oldsubparagraph\subparagraph
  \renewcommand{\subparagraph}{
    \@ifstar
      \xxxSubParagraphStar
      \xxxSubParagraphNoStar
  }
  \newcommand{\xxxSubParagraphStar}[1]{\oldsubparagraph*{#1}\mbox{}}
  \newcommand{\xxxSubParagraphNoStar}[1]{\oldsubparagraph{#1}\mbox{}}
\fi
\makeatother

\usepackage{longtable,booktabs,array}
\usepackage{calc} 
\usepackage{etoolbox}
\makeatletter
\patchcmd\longtable{\par}{\if@noskipsec\mbox{}\fi\par}{}{}
\makeatother
\IfFileExists{footnotehyper.sty}{\usepackage{footnotehyper}}{\usepackage{footnote}}
\makesavenoteenv{longtable}
\usepackage{graphicx}
\makeatletter
\def\maxwidth{\ifdim\Gin@nat@width>\linewidth\linewidth\else\Gin@nat@width\fi}
\def\maxheight{\ifdim\Gin@nat@height>\textheight\textheight\else\Gin@nat@height\fi}
\makeatother
\setkeys{Gin}{width=\maxwidth,height=\maxheight,keepaspectratio}
\makeatletter
\def\fps@figure{htbp}
\makeatother

\makeatletter
\@ifpackageloaded{caption}{}{\usepackage{caption}}
\AtBeginDocument{%
\ifdefined\contentsname
  \renewcommand*\contentsname{Table of contents}
\else
  \newcommand\contentsname{Table of contents}
\fi
\ifdefined\listfigurename
  \renewcommand*\listfigurename{List of Figures}
\else
  \newcommand\listfigurename{List of Figures}
\fi
\ifdefined\listtablename
  \renewcommand*\listtablename{List of Tables}
\else
  \newcommand\listtablename{List of Tables}
\fi
\ifdefined\figurename
  \renewcommand*\figurename{Figure}
\else
  \newcommand\figurename{Figure}
\fi
\ifdefined\tablename
  \renewcommand*\tablename{Table}
\else
  \newcommand\tablename{Table}
\fi
}
\@ifpackageloaded{float}{}{\usepackage{float}}
\floatstyle{ruled}
\@ifundefined{c@chapter}{\newfloat{codelisting}{h}{lop}}{\newfloat{codelisting}{h}{lop}[chapter]}
\floatname{codelisting}{Listing}

\makeatother
\makeatletter
\@ifpackageloaded{caption}{}{\usepackage{caption}}
\@ifpackageloaded{subcaption}{}{\usepackage{subcaption}}
\makeatother

\ifLuaTeX
  \usepackage{selnolig}  
\fi
\usepackage[]{natbib}
\usepackage{bookmark}

\IfFileExists{xurl.sty}{\usepackage{xurl}}{} 
\hypersetup{
  pdftitle={Title},
  pdfauthor={Author 1; Author 2},
  pdfkeywords={3 to 6 keywords, that do not appear in the title},
  colorlinks=true,
  linkcolor={blue},
  filecolor={Maroon},
  citecolor={Blue},
  urlcolor={Blue},
  pdfcreator={LaTeX via pandoc}}

\newcommand{\anon}{1}

\usepackage{mathtools}          
\usepackage{empheq}             
\usepackage{bm}
\usepackage{bbm, dsfont}                
\usepackage{derivative}
\allowdisplaybreaks
 
\usepackage{algorithm}
\usepackage[noend]{algpseudocode}
\algnewcommand{\algorithmicand}{\textbf{ and }}
\algnewcommand{\algorithmicor}{\textbf{ or }}
\algnewcommand{\OR}{\algorithmicor}
\algnewcommand{\AND}{\algorithmicand}
\algnewcommand{\IfThenElse}[3]{%
  \State \algorithmicif\ #1\ \algorithmicthen\ #2\ \State \algorithmicelse\ #3}
\algnewcommand{\IfThen}[2]{%
  \State \algorithmicif\ #1\ \algorithmicthen\ #2}
\algrenewcommand\algorithmicrequire{\textbf{Input:}}
\algrenewcommand\algorithmicensure{\textbf{Output:}}
 
\usepackage{multirow}
\usepackage{enumerate}
\usepackage{lastpage}

\usepackage{setspace}   
 
\newtheorem{theorem}{Theorem}[section]

\newtheorem{proposition}[theorem]{Proposition}

\newtheorem{definition}[theorem]{Definition}

\newtheorem{remark}[theorem]{Remark}

\DeclarePairedDelimiter\abs{\lvert}{\rvert}

\newcommand{\R}{\mathbb{R}}
\newcommand{\ev}{\mathbb{E}}
\DeclareMathOperator{\Var}{Var}
\DeclareMathOperator{\Corr}{Corr}

\newcommand{\Z}{\mathbb{Z}}

\DeclareMathOperator*{\argmin}{arg\,min}

\newcommand{\tldq}{\tilde{q}}
\newcommand{\tldz}{\tilde{\z}}
\newcommand{\tldZ}{\tilde{\mathbf{Z}}}
\newcommand{\bt}{\boldsymbol{\theta}}
\newcommand{\vt}{\boldsymbol{\vartheta}}
\newcommand{\x}{\boldsymbol{x}}
\newcommand{\y}{\boldsymbol{y}}
\newcommand{\pss}{\boldsymbol{\psi}}
\newcommand{\z}{\mathbf{z}}
\newcommand{\uu}{\mathbf{u}}

\DeclarePairedDelimiterX{\infdivx}[2]{(}{)}{#1\;\delimsize\|\;#2}
\newcommand{\kl}{D_\textrm{KL}\infdivx}
 
\newcommand{\floor}[1]{\left\lfloor #1 \right\rfloor}

\newcommand{\defeq}{\vcentcolon=}

\begin{document}

\def\spacingset#1{\renewcommand{\baselinestretch}%
{#1}\small\normalsize} \spacingset{1}


\if1\anon
{
  \title{\bf Simulation-based inference via telescoping ratio estimation
for trawl processes}
\author{
Dan Leonte$^{1}$\thanks{dan.leonte@kaust.edu.sa}, 
Rapha\"el Huser$^{1}$,
 Almut E. D. Veraart$^{2}$ \\[0.5em]
$^{1}$Computer, Electrical and Mathematical Sciences
and Engineering Division,\\ King Abdullah University of Science and Technology, \\
Thuwal, Saudi Arabia \\[0.5em]
$^{2}$Department of Mathematics, Imperial College London, \\
London, United Kingdom
}
  \maketitle
} \fi

\if0\anon
{
  \bigskip
  \bigskip
  \bigskip
  \begin{center}
    {\LARGE\bf Simulation-based inference via telescoping ratio estimation
for trawl processes}
\end{center}
  \medskip
} \fi

\bigskip
\begin{abstract}
The growing availability of complex datasets has increased interest in temporal stochastic processes capable of capturing stylized facts such as non-Gaussian marginals, long memory, and non-Markovian dynamics. While such models are often easy to simulate, parameter estimation remains challenging. Simulation-based inference (SBI) offers a promising way forward, but existing approaches typically require large training datasets or complex architectures and often yield confidence (credible) regions that fail to attain nominal coverage, questioning the reliability of estimates for the very features that make these models appealing. To address these challenges, we propose a fast, accurate, and sample-efficient SBI framework for amortized posterior inference in intractable stochastic processes. Our method relies on two steps: first, we learn the posterior density by decomposing it sequentially across parameter dimensions; second, we employ Chebyshev polynomial approximations to efficiently generate independent posterior samples, enabling accurate inference even when Markov chain Monte Carlo methods mix poorly. Moreover, we develop novel diagnostic tools and post-hoc calibration techniques that improve performance and allow reusing the trained inference network with time series of varying lengths, further amortizing training costs. We demonstrate the effectiveness of the method on trawl processes, a flexible class of infinitely divisible models that generalize univariate Gaussian processes, applied to energy data.\end{abstract}

\noindent%
\vfill

\newpage
\spacingset{1.8}

\section{Introduction}

Many classes of stochastic processes are straightforward to simulate from but have likelihood functions that are either difficult or impossible to evaluate. This asymmetry is increasingly common with processes that exhibit complex dependence structures or model high-dimensional observations, making classical likelihood-based inference computationally prohibitive. 
When efficient simulators are available, neural simulation-based inference (SBI) provides a viable alternative by replacing direct likelihood evaluation with inference procedures trained on simulated parameter-data pairs $(\bt,\x)$. Once trained, such procedures can be amortized across observations, making neural SBI especially attractive when inference must be repeated for many datasets generated by the same model.

SBI has traditionally focused on variants of approximate Bayesian computation (ABC). 
Given an observation $\x_{\mathrm{o}}$, ABC generates many parameter-data pairs $(\x,\bt)$ and only retains those $\bt$ for which $\x$ closely matches $\x_{\mathrm{o}}$, i.e., $d\left(\x,\x_{\mathrm{o}}\right) < \epsilon$ for some metric $d(\cdot,\cdot)$ and error tolerance $\epsilon > 0$. Several works attempt to improve the impractically low ABC acceptance rates by comparing informative summary statistics rather than the raw data \citep{blum2013comparative} or by embedding ABC within Markov chain Monte Carlo (MCMC) \citep{marjoram2003markov}, but the methodology remains sensitive to the choice of metric $d$ and tolerance $\epsilon$ \citep{robert2011lack}, and is generally computationally inefficient. By contrast, modern SBI has increasingly relied on neural inference techniques, which use simulations $(\x,\bt)$ to train a neural network that directly approximates either (1) the likelihood $p(\x \mid \bt)$, exactly \citep{papamakarios_2017_neurips,papamakarios_2019_JMLR} or up to a constant \citep{hinton_2002_ebm_CD}, (2) the 
posterior $p(\bt \mid \x)$ \citep{radev_2020}, or (3) point summaries thereof \citep{matt_american_statistician,matt_jcgs,Richards_et_al}. Once trained, the network can be evaluated on new observations without retraining, hence amortizing the initial training cost; see \cite{zammit_mangion_annual_review_2025} for a review of amortized inference with neural methods.

Among neural SBI methods, we focus on neural ratio estimation (NRE), which avoids the complex architectures required by flow-based models and turns intractable density-ratio estimation into a binary classification problem. A neural network is trained to distinguish between samples from the joint distribution $p(\x,\bt)=p(\x\mid\bt)p(\bt)$ and samples drawn independently from the product of marginals $p(\x)p(\bt)$, where $p(\bt)$ is a sampling distribution chosen to ensure good parameter-space coverage during training and $p(\x)=\int p(\x\mid\bt)p(\bt)\,\mathrm{d}\bt$ is the induced marginal data distribution. Crucially, $p(\bt)$ serves only as a sampling distribution for generating training data, 
and a different distribution can be used as prior for Baysian inference.
The odds of the optimal classifier recovers the ratio
$ r(\x,\bt)=\frac{p(\x,\bt)}{p(\x)p(\bt)}=\frac{p(\x\mid\bt)}{p(\x)},
$
which is proportional to the full likelihood for fixed $\x$, enabling both frequentist and Bayesian inference.

Despite this appealing reduction to classification, standard NRE becomes difficult precisely in the settings where SBI is most useful. For example, as the dimensionality of $\bt$ or the amount of information contained in $\x$ increases, the joint density $p(\x,\bt)$ becomes highly concentrated relative to the product of marginals $p(\x)p(\bt)$. Consequently, the ratio $r(\x,\bt)$ takes extremely large values in a narrow region of parameter space and is nearly zero elsewhere, and the classifier struggles to learn meaningful level sets. While it is well known that NRE can produce posterior approximations with poorly calibrated credible regions \citep{hermans2022trust}, we uncover an additional critical failure: classical NRE can also fail to accurately locate the likelihood mode itself. This dual failure, affecting both uncertainty quantification and point estimation, represents a fundamental breakdown of the standard approach in challenging inference problems.

\textbf{Contributions:} To address these challenges, we make four key contributions that advance NRE-based SBI along several fronts: (i) improved sample efficiency during training, (ii) reduced computational cost during inference, (iii) diagnostic tools to detect errors in the learnt likelihood, and (iv) post-training calibration techniques that correct these errors and enable additional amortization. The proposed methodology is model-agnostic, requiring only a finite-dimensional parameterization of the model and access to a 
simulator.

In our first contribution (i), we adapt telescoping ratio estimation \citep[TRE;][]{rhodes_tre} to the SBI setting by decomposing the likelihood ratio as a product of $m$ simpler, one-dimensional conditional ratios,
$
 r(\x,\bt)=\prod_{i=1}^m r_i\left(\x,\bt^{1:i}\right),
$
where the term $r_i$ depends on $\bt\in\R^m$ only through its first $i$ components, denoted $\bt^{1:i}$. This decomposition simplifies the learning task, dramatically improving training efficiency. We formally show that this setup requires exponentially fewer samples $(\x,\bt)$ to reach the same approximation quality as standard NRE. In our second contribution (ii), we introduce a novel, MCMC-free, GPU-friendly sequential sampling approach that builds on the proposed TRE decomposition and significantly accelerates posterior inference compared with state-of-the-art schemes such as the No U-Turn Sampler (NUTS). By accurately approximating each of the one-dimensional conditional densities $\{p(\theta^i\mid\x,\bt^{1:i-1})\}_{i=1}^m$ using Chebyshev polynomials \citep{olver2013fast}, which converge uniformly on compacts, we enable efficient inverse transform sampling with a computational cost that is linear in the dimensionality $m$ of the parameter $\bt$. While this adds a second approximation layer on top of TRE, it crucially decouples expensive neural-network evaluations from downstream tasks such as posterior sampling and the computation of highest posterior density regions. We also generalize this sequential approach to classifiers that model two components of $\bt$ jointly. In our third contribution (iii), we introduce new component-wise diagnostics that extend existing SBI validation methods and make them computationally efficient using the proposed Chebyshev approximations. Finally, in our fourth contribution (iv), we show that even after decomposing the learning problem into simpler tasks through TRE, the classifiers still benefit from post-training calibration, since calibration is a global property that cannot be reliably enforced from mini-batches during stochastic gradient descent. We further show that, in the time series context, calibration can be exploited to amortize the SBI estimator across different time series lengths.

We study these methodological developments using trawl processes \citep{ole_brazilian} as a challenging test case, for three main reasons. First, trawl processes form a flexible class of possibly 
non-Markovian 
 processes within the framework of Ambit stochastics \citep{ambit_book}, extending continuous-time Gaussian processes to the infinitely divisible setting. They can capture asymmetric and heavy-tailed marginal distributions together with a wide range of serial dependence structures, including both short- and long-memory behavior. Second, their statistical properties are fully characterized by their marginal distribution and autocorrelation function, both of which are available in closed form and easy to parameterize. Third, while their likelihood function is generally intractable \citep{leonte2023likelihood}, efficient simulation algorithms are available \citep{leonte2024simulation}.

We illustrate the practical utility of our methodology and the benefits of amortization in a simulation study on trawl processes and an application to energy demand data. In doing so, we also enable the first full-posterior inference for trawl processes. 

\textbf{Paper outline:} 
Section~\ref{sec-meth} introduces our proposed SBI methodology based on the novel TRE decomposition and the use of Chebyshev polynomials. Section~\ref{section:simulation_study} demonstrates our SBI methodology in a simulation study with trawl processes. Section~\ref{section:application} applies the methodology to real-world energy demand data. Section~\ref{section:conclusion} concludes with implications for the broader SBI literature and directions for future research. Additional details and results are available in the Supplementary Material.

\section{Simulation-based inference methodology}\label{sec-meth}
In this section, we first 
recall background on the NRE framework for SBI 
and discuss its main limitations. We then introduce our novel SBI methodology, and its theoretical foundations, which combines telescoping ratio estimation (TRE) for improved training efficiency and polynomial approximations in the basis of Chebyshev polynomials for fast MCMC-free posterior sampling. We also discuss post-training calibration strategies to remedy the approximation error and enable amortization over the series length in the context of time series data. 
Finally, we propose novel tests and diagnostics to assess the quality of the learnt likelihood ratio. 
\subsection{NRE: learning likelihood ratios through classification}\label{subsection:NRE_learning_likelihood_through_classification}
The idea of training a probabilistic classifier to learn likelihood ratios dates back to \cite{bickel2007discriminative}, 
 who noted that maximum likelihood estimation is not consistent under covariate shift, i.e., when the testing and training data have different densities $q$ and $\tldq$. To account for this, the ratio $\z \mapsto r(\z) \defeq q(\z) / \tldq(\z)$ must be approximated for all $\z \in \mathcal{Z}$. The authors considered a binary classifier $c \colon \mathcal{Z} \to [0,1]$ that distinguishes between samples from $\mathbf{Z} \sim p(\z \mid Y=1) \defeq  q(\z)$ and $\tldZ \sim p(\tldz \mid Y=0) \defeq \tldq(\tldz)$, where the label $Y$ is balanced a priori, i.e., $p(Y=1) = p(Y=0) = 0.5$. We use $\z$ to denote both an input to the classifier $c$ and a realization from $\mathbf{Z}\sim q$, with the meaning being clear from the context. Let $c^{*}$ be the Bayes optimal classifier with respect to the binary cross-entropy loss (BCE), i.e., 
\begin{equation}
c^{*} \defeq \argmin_{c} \, \ev\left[-\log{c(\mathbf{Z})} - \log{(1-c(\tilde{\mathbf{Z}}))} \right].\label{eq:theretical_bce_with_z}
\end{equation}
By a calculus of variations argument \citep[Proposition 2]{Cranmer2015ApproximatingLR}
, it follows that
\begin{equation}
    c^{*}(\z) = \frac{q(\z)}{q(\z) + \tldq(\z)}.
\label{eq:optimal_classifier}
\end{equation}
Assuming the Bayes optimal classifier is available, the likelihood ratio can be recovered as
\begin{equation*}
r(\z) \defeq \frac{q(\z)}{\tldq(\z)} = \frac{q(\z) / \left({q(\z) + \tldq(\z)}\right)}{\tldq(\z) / \left({q(\z) + \tldq(\z)}\right)}
= \frac{c^{*}(\z)}{1-c^{*}(\z)}.
\end{equation*}
In practice, neither the BCE loss from \eqref{eq:theretical_bce_with_z} nor $c^{*}$ are tractable. In lieu, consider a flexible set of parametric classifiers $\{c_{\pss} : \pss \in \Psi\}$ satisfying a universal approximation theorem, e.g., based on neural networks, and minimize the Monte Carlo approximation of the BCE loss
\begin{equation}
\frac{1}{N}\left[-\sum_{i=1}^{N}\log{\left(c_{\pss}(\z_i)\right)} - \sum_{i=1}^N\log{\left(1-c_{\pss}(\tldz_i)\right)} \right],\label{eq:BCE_approx_from_samples}
\end{equation} 
where $\left\{\z_i \right\}_{i=1}^N \stackrel{\text{iid}}{\sim} q$ and $\left\{\tldz_i \right\}_{i=1}^N \stackrel{\text{iid}}{\sim} \tldq$, to obtain a minimizer $\hat{\pss}$ and the corresponding approximations  $\hat{c} \defeq c_{\hat{\pss}}$ of $c^{*}$ and $\hat{r}$ of $r$; see Algorithm \ref{algo:general_nre} in Section~\ref{subsection:TRE_training_procedure}  of the Supplementary Material for pseudocode. \cite{Cranmer2015ApproximatingLR} and \cite{amortized_hermans_2020} pioneer the above classification-based approach within SBI, when fast samplers are available. Specifically, let $\x \in \mathcal{X} \subset \R^k$ be a realization of a stochastic process parameterized by $\bt \in \Theta \subset \R^m$, in our case a trawl process, and let $\z = (\x,\bt) \in \mathcal{Z} = \mathcal{X} \times \Theta \subset \R^{k+m}$. For ease of notation, we write $p(\x \mid \bt)$ for the likelihood function in both the frequentist and Bayesian cases. Consider a sampling density $p(\bt)$ on $\Theta$ and the induced joint, marginal, and product of marginal densities $q(\x,\bt) \defeq p(\x,\bt ) = p(\x \mid \bt) p(\bt)$, $p(\x) = \int p(\x,\bt) \mathrm{d} \bt$, and $\tldq(\x,\bt) \defeq  p(\x)p(\bt)$, respectively. A classifier $\hat{c}$ trained to distinguish between samples from $q$ and $\tldq$ yields an approximation  $(\x,\bt) \mapsto \hat{r}(\x,\bt)$ of  
\begin{equation*}
 r(\x,\bt) =  \frac{q(\x,\bt)}{\tldq(\x,\bt)} = \frac{p(\x,\bt)}{p(\x)p(\bt)} = \frac{p(\x \mid\bt)}{p(\x)} \ \propto  \ p(\x\mid\bt) \text{ for fixed } \x. 
\end{equation*} 
See Algorithm \ref{algo:specialised_nre} in Section~\ref{subsection:TRE_training_procedure} of the Supplementary Material for 
pseudocode and how samples from $p(\x)p(\bt)$ can be generated by marginalizing samples from $p(\x,\bt)$. Thus, NRE facilitates both frequentist and Bayesian inference, as $\hat{r}$ approximates the likelihood up to a normalizing constant, while the posterior is approximated via Bayes' rule as $p(\bt \mid \x) = r(\x,\bt) p(\bt) \approx \hat{r}(\x,\bt) p(\bt)$ if $p(\bt)$ is also used as the prior. The prior used for inference does not have to match the sampling distribution used during training. In this case, the posterior is available up to a constant. Notably, the approximation $r \approx \hat{r}$ is valid only in regions of the parameter space well covered by training samples, 
hence the sampling density $p(\bt)$ is often chosen as a product of uniform distributions. 

However, unlike the Bayes optimal classifier $c^{*}$, minimizers based on finite sample approximations of the BCE loss \eqref{eq:BCE_approx_from_samples}, or of related losses used with other density estimation methods, are known to be ill-calibrated \citep{hermans2022trust}. This miscalibration is a fundamental issue for SBI that affects confidence (credible) regions. Several partial remedies have been proposed for NRE: \cite{mukhoti2020calibrating} and \cite{delaunoy2023balancing} amend the BCE loss in an attempt to obtain conservative classifiers, which is the preferred mode of failure, yet this does not fully address miscalibration; \cite{falkiewicz2023calibrating} differentiably incorporate posterior coverage in the loss function, but doing so requires posterior samples from $p(\bt \mid \x)$ during training---a strategy that becomes computationally infeasible when the parameter space $\Theta$ exceeds two or three dimensions. Additional methods target truncations or marginalized versions of the likelihood function \citep{miller2021truncated}, which can fail to capture interactions among the components of $\bt$. Finally, applying post-training calibration via Platt-scaling or similar, as suggested in \cite{Cranmer2015ApproximatingLR}, typically fails when NRE is employed within SBI. This failure is symptomatic of a deeper issue in the framework itself: the more complicated the stochastic process $X$ 
and the higher the dimensionality of $\bt$, the `further apart' $q(\x,\bt) = p(\x,\bt)$ and $\tldq(\x,\bt) = p(\x) p(\bt)$ become, and the easier it is for a classifier to trivially separate samples from $q$ and $\tldq$. Approximating the ratio of very different densities is a well-documented challenge in the literature \citep{
NIPS2011_d1f255a3}. \cite{persi2018sample} and \cite{ mcallester2020formal} prove this task is inherently sample-inefficient, in the sense that a sample size that is exponential in the Kullback--Leibler (KL) divergence $\kl{q}{\tldq}$ is necessary for accurate estimation. 
With insufficient samples, classifiers tend to degenerate and only output values that are approximately $0$ and $1$, with many models achieving low BCE loss while remaining poorly calibrated. This is a significant issue, as NRE is employed to accurately recover the level sets of the likelihood ratio, which in turn define the contours of the likelihood $p(\x \mid \bt)$ and the posterior $p(\bt \mid \x)$. 

\subsection{TRE: a novel divide-and-conquer SBI technique} \label{section:meth_contr_to_nre}
To address the above issue, we introduce telescoping ratio estimation (TRE) to the 
SBI framework. TRE mitigates the inherent difficulty of ratio estimation by decomposing the target likelihood ratio into a product of simpler ratios, each involving pairs of distributions that are closer in terms of KL divergences. This decomposition allows the corresponding ratios to be learned separately using multiple low-complexity classifiers, whose outputs are then multiplied to recover the target likelihood ratio. For an optimal choice of interpolants, this novel approach requires exponentially fewer samples compared to NRE.

Formally, we introduce interpolating distributions $q_0,\ldots,q_m$, such that $q_0 = \tldq$ and $q_m = q$, and learn the likelihood-ratio by training $m$ classifiers and multiplying the corresponding ratios. 
Writing $\bt = (\theta^1,\ldots,\theta^m) \in \Theta \subset \R^m$ and $\bt^{i:j}=(\theta^i,\ldots,\theta^j)$ 
for $1\leq i\leq j\leq m$, define
\begin{equation*} \vspace{-0.25cm} 
q_i(\x,\bt) 
=p(\x,\bt^{1:i} ) p(\bt^{i+1:m} \mid \bt^{1:i}), \qquad i = 1, \ldots, m-1,  \label{eq:dimensionwise_mixing_densities} \vspace{-0.25cm} 
\end{equation*}
and $q_0(\x,\bt) =  p(\x)p(\bt)$, $q_m(\x,\bt) =  p(\x,\bt)$. Furthermore, define the density ratios    
\begin{equation*}\vspace{-0.15cm} 
    r_i(\x,\bt^{1:i}) \defeq  \frac{q_i(\x,\bt)}{q_{i-1}(\x,\bt)} = 
\frac{p(\x,\bt^{1:i}) p(\bt^{i+1:m} \mid \bt^{1:i})}{p(\x,\bt^{1:i-1} ) p(\bt^{i:m} \mid \bt^{1:i-1})} = \frac{p(\theta^{i} \mid \x,\bt^{1:i-1})}{p(\theta^{i} \mid \bt^{1:i-1})}, \qquad i=1,\ldots,m,\vspace{-0.15cm} 
    \end{equation*}
with the convention that $\bt^{1:0}=\bt^{m+1:m}=\emptyset$, and note $r_i$ depends on $\bt$ only through its first $i$ components $\bt^{1:i}$. The desired ratio $r(\x,\bt) = p(\x,\bt) / \left(p(\x)p(\bt)\right)$ can then be obtained as $r(\x,\bt) = \prod_{i=1}^m r_i(\x,\bt^{1:i})$. 
One of the main benefits is that the density ratios $r_i$ can be approximated separately, by training $m$ classifiers to distinguish between samples from $q_{i}$ and $q_{i-1}$, for $i =1 ,\ldots,m$. Each of these intermediate ratios are flatter and less sharply peaked than the full ratio, making the corresponding classification problems better conditioned than the original one. The choice of the interpolating distributions $\{q_i\}_{i=0}^m$ is crucial and directly linked to training efficiency, as revealed by Theorem \ref{thm:sample_efficiency}. 
\begin{theorem}[Sample efficiency]\label{thm:sample_efficiency}
Let $q_0(\x,\bt) = p(\x)p(\bt)$, $q_m(\x,\bt) = p(\x,\bt)$, and $q_i(\x,\bt) =p(\x,\bt^{1:i}) p(\bt^{i+1:m} \mid \bt^{1:i})$ 
for $i = 1,\ldots, m-1$, as above. Then, 
\begin{equation*}
\kl{p(\x,\bt)}{p(\x)p(\bt)} = \kl{q_m}{q_0} = \sum_{i=1}^m \kl{q_i}{q_{i-1}}.
\end{equation*}
\label{prop:kl_decomposition}
\end{theorem}\vspace{-1.25cm} 
See Section~\ref{appendix:subsection_tre}  of the Supplementary Material for the proof. While NRE requires exponentially many samples in $\kl{q_m}{q_0}$ to accurately represent the BCE loss \citep{persi2018sample, mcallester2020formal}, the $i^\text{th}$ of the $m$ TRE classifiers only requires exponentially many samples in $\mathcal{S}_i \defeq \kl{q_i}{q_{i-1}}$, which implies that the simulation budget can be reduced from $\exp{(\sum_{i=1}^m S_i)}$ in NRE to $\sum_{i=1}^m \exp{(S_i)}$ in TRE. Further, each simulated pair $(\x,\bt)$ contains all lower dimensional pairs $(\x,\bt^{1:i})$ for $i=1,\ldots,m$, hence the same simulations can be used to train all $m$ classifiers. This reuse couples the empirical losses across classifiers but leaves each estimator unbiased, with the sampled $(\x,\bt)$ effectively acting as ``common random numbers''. Consequently, for a fixed accuracy, the required number of simulator calls grows exponentially in $\max_{i = 1\ldots, n} S_i$ for TRE instead of $\sum_{i=1}^m S_i$ for NRE. If the values $\mathcal{S}_i$ are equal, this reduces the exponent by a factor of $1/m$. Importantly, the quantities $\mathcal{S}_i$ are not merely theoretical: they can be estimated and monitored during training directly from the classifier outputs at no extra computational cost since $\mathcal{S}_i = \kl{q_i}{q_{i-1}} \approx \ev_{q_i(\x,\bt)}\left[\log{\hat{r}}_i(\x,\bt)\right]$. Moreover, 
the $\mathcal{S}_i$ values can inform the allocation of batch sizes and compute across 
classifiers, e.g., bigger batches and longer training can be assigned to classifiers with larger $\mathcal{S}_i$ (i.e., more difficult learning tasks).
\begin{remark}
    The order in which parameters are learned affects the resulting KL divergences $\mathcal{S}_i, \, i=1,\ldots,m$, hence the difficulty of each classification problem and overall training efficiency. Permuting the order in which parameters are learnt amounts to modifying the $\mathcal{S}_i$ values by conditional mutual information values (see Section~\ref{subsec:TRE_ordering_SM} of the Supplementary Material for details), while keeping $\sum_{i=1}^m \mathcal{S}_i$ constant. Although theoretically there is an optimal coordinate ordering, in practice, we do not notice significant differences in performance. 
    \end{remark}
The same sequential decomposition allows to control the propagation of individual approximation errors from each learnt 
conditionals to the full posterior. Given the learnt ratios $\hat{r}_i$, the conditional densities 
\begin{equation*}
    \hat{p}(\theta^i \mid \x, \bt^{1:i-1}) = p(\theta^i \mid \bt^{1:i-1}) \hat{r}_i(\x,\bt^{1:i}), \qquad i=1,\ldots,m,
\end{equation*}
induce the approximate posterior $\hat{p}(\bt \mid \x) = \prod_{i=1}^m \hat{p}(\theta^i \mid x, \bt^{1:i-1})$. Proposition~\ref{thm:approx_error} follows.
\begin{proposition}[Posterior error accumulation]\label{thm:approx_error}
For any fixed $\x$,
\begin{align*}
\kl{p(\bt\mid\x)}{\hat p(\bt\mid\x)}
&=
\sum_{i=1}^m
\ev_{p(\bt^{1:i-1}\mid\x)}
\left[
\kl{p(\theta^i\mid\x,\bt^{1:i-1})}
{\hat p_i(\theta^i\mid\x,\bt^{1:i-1})}
\right].
\end{align*}
\end{proposition}
The result can be proved by repeated use of the chain rule for differential entropy~\cite[Theorem~2.5.3]{cover_thomas}. Thus, approximation errors across parameter-components accumulate additively rather than compounding through the product of ratios. 
\begin{remark}[Blockwise extension]
    The componentwise construction above is not restricted to learning one parameter at a time. More generally, we can partition $\bt\in\R^m$ into $l<m$ subvectors as $\bt=(\bt^{(1)},\dots,\bt^{(l)})$, and learn the block‐conditional densities ${p(\bt^{(j)} \mid \x,\bt^{(1:j-1)})}$, $j=1,\ldots,l$. 
    Theorem~\ref{thm:sample_efficiency} and Proposition~\ref{thm:approx_error} extend directly to this setting. This makes it possible to combine parameters associated with small intermediate KL divergences into single blocks, reducing the total number of classifiers while preserving the theoretical guarantees.
\end{remark}
\par
\textbf{Related work:} The closest work to our proposed 
methodology is \cite{rhodes_tre}, who also adapt principles from simulated annealing \citep{Kirkpatrick_simulated_annealing, neal2001annealed} 
to the neural setup. Their construction targets unconditional density estimation of a single, fixed dataset via data-to-noise waymarks, whereas we develop the telescoping principle for amortized SBI, with interpolants defined across parameter dimensions. In other words, our method is used for likelihood ratio estimation conditional on (potentially high-dimensional) data; it can be applied without re-training to any 
dataset from the same statistical model and its success relies on suitably defined interpolating classifiers. Our construction also addresses several 
limitations raised by \cite{rhodes_tre}: optimality of interpolants, the train--inference mismatch, and the lack of fast sampling. Specifically, our interpolants provide an exact sample-efficiency characterization and are optimal from an information-theoretic viewpoint. 
Theorem~\ref{thm:sample_efficiency} indeed shows that the KL divergences sum up, without overhead, to the global KL divergence (between the target joint density and the product of marginals), so the
interpolants trace a geodesic-type path with respect to the KL divergence, thereby
resolving the theoretical question listed as open in the conclusion of \cite{rhodes_tre}.
Furthermore, in \citet{rhodes_tre}'s formulation, classifiers are trained on consecutive interpolants and evaluated at a
common input at inference time, which amounts to evaluating the trained neural network on samples that are out of the distribution of the training data. \cite{rhodes_tre} acknowledge this limitation in their Section~3.2, and it motivated
the follow-up multinomial-classification construction of
\cite{srivastava2023tmlr}. In our formulation, posterior sampling instead proceeds sequentially: we first evaluate $\hat{r}_1(\x,\theta^1)$ to obtain and sample from $\hat{p}(\theta^1\mid\x)$; conditional on the resulting draw $\theta^1$, we evaluate $\hat{r}_2(\x,\bt^{1:2})$ to sample from $\hat p(\theta^2\mid\x,\theta^1)$, and so on. Hence, each ratio is evaluated conditional on parameter values generated by the preceding learnt conditionals, avoiding the train--inference mismatch by construction. Finally, the same sequential factorization 
gives rise to the fast MCMC-free posterior sampler introduced in the next section.
\subsection{MCMC-free posterior sampling via Chebyshev polynomials}\label{subsection:mcmc_posterior_sampling} 
Fast posterior sampling is crucial both for practical applications and for the coverage checks described in Section~\ref{subsection:posterior_checks}. These checks indeed require to repeatedly simulate parameters $\bt$, generate data $\x$ from the model, and draw thousands of posterior samples for each simulated pair $(\bt,\x)$. More broadly, efficient posterior sampling is a central requirement in Bayesian inference, as many inferential tasks rely on repeated posterior sampling and become computationally prohibitive without scalable sampling.

In practice, obtaining satisfactory effective sample sizes would typically require adaptive samplers such as No U-Turn Sampler (NUTS), which tune not only the proposal covariance and step size, but also the number of leapfrog steps. Even such advanced MCMC schemes can be inefficient in our setting: their adaptive nature prevents efficient GPU vectorization and forces Markov chains to be run sequentially. This motivates the development of a novel, MCMC-free approach.

Having gained access to the one-dimensional conditional densities $p(\theta^i \mid \x, \bt^{1:i-1})$ through the ratios $r_i$ for $i =1 ,\ldots,m$, we discuss the efficient generation of posterior samples sequentially in the components of $\bt$. Instead of MCMC, we propose to use an efficient inversion sampling algorithm, by constructing approximations of the conditional cumulative distribution functions (CDFs) 
$F\left(\theta^i \mid \x, \bt^{1:i-1}\right)$ which are simple to evaluate and then invert by bisection \citep{olver2013fast}. A detailed account is provided in Section~\ref{suppl:subsection_chebyshev} of the Supplementary Material, and we only summarize here the 
main steps. First, we evaluate $\theta^i\mapsto\hat {p}(\theta^i\mid \x,\bt^{1:i-1})$ at $N$ suitable knots, called Chebyshev knots, and determine the coefficients of the resulting approximation in the basis of Chebyshev polynomials using a discrete cosine transform that only requires $\mathcal{O}(N\log N)$ operations. Second, an approximation of the target CDF can be obtained in closed form 
and is evaluated stably via Clenshaw's algorithm \citep{clenshaw1955note} using $\mathcal{O}(N)$ operations. Finally, independent draws are obtained by inverting this CDF by bisection, which 
requires up to $24$ (respectively, $53$) steps to converge to $\textrm{float32}$  (respectively, $\textrm{float64}$) precision. The key aspect here is that all these cheap intermediate operations replace the more expensive neural network calls.

This construction is also well suited for GPU execution: the neural-network evaluations at the Chebyshev knots, coefficient computation, integration and CDF inversion are all vectorized, and the approximation converges geometrically for analytic functions \citep{olver2013fast}; for smooth, $s$-times continuously-differentiable but non-analytic functions, the approximation error decays like $\mathcal{O}(N^{-s})$. In practice, we find that a modest number of $N=128$ knots is sufficient to make the Chebyshev approximation error negligible even when non-smooth activation functions are used, e.g., ReLU. For fixed $N$, the resulting complexity scales linearly with the parameter dimension $m$. Finally, this approach allows us to introduce novel, computationally efficient checks to individually test each component classifier, as 
detailed in Section~\ref{subsection:posterior_checks}. Since the generated samples are independent, we avoid thinning procedures that are often necessary with MCMC samples for simulation-based calibration diagnostics  
\citep{talts2018validating}. 
\subsection{Calibration analysis and further amortization}\label{subsection:postcalibration}
\subsubsection{Potential reasons for miscalibration and remedies} \label{subsection:potential_reasons}
We say that a binary classifier $c : \mathcal{Z} \to [0,1]$ is calibrated if the following holds:
\begin{equation*}
    \mathds{P}\left(Y=1\mid c(\mathbf{Z})\right) = c(\mathbf{Z}) \, \text{ almost surely  (a.s.) over pairs } (\mathbf{Z},Y), 
\end{equation*}
where $Y \in \{0,1\}$ denotes the true label of data $\mathbf{Z}$. Intuitively, among all inputs $\z$ for which $c(\z) = 0.75$ say, 
$75\%$ should have label $Y=1$. Nevertheless, neural network-based classifiers, including NREs
, are often miscalibrated \citep{on_calibration_of_modern_nn, hermans2022trust}. Miscalibration is typically quantified using the expected calibration error (ECE), $\ev\left[\left|\mathds{P}\left(Y =1 \mid c(\mathbf{Z}) \right)-c(\mathbf{Z}) \right|\right]$, which measures the average discrepancy between the predicted confidence $c(\mathbf{Z})$ and the accuracy $\mathds{P}\left(Y =1 \mid c(\mathbf{Z})\right)$, and can be 
estimated from samples, 
see Section~\ref{SM:ECE} of the Supplementary Material. Empirical studies \citep{caruana2004predicting, on_calibration_of_modern_nn, minderer2021revisiting_calibration} report conflicting results, with calibration varying based on regularization, architecture and model capacity. We identify two main avenues for improvement and 
leverage the rich literature on post-hoc calibration of probabilistic classifiers to ensure that the level curves of the likelihood (posterior) are calibrated. 

The first issue arises from training with a fixed, finite dataset. If samples from the joint and product of marginal distributions are (nearly) separable, and provided the classifier is sufficiently expressive, continued optimization of the empirical BCE loss can drive outputs values to degenerate towards $0$ and $1$. This can occur even when the underlying distributions overlap, because a finite dataset may miss entirely the rarer observations in regions where the two classes are difficult to distinguish, unlike the theoretical BCE objective, which is minimized at the optimal decision function from \eqref{eq:optimal_classifier}. This is formally analyzed by \cite{soudry2018implicit} for linearly separable data and is particularly problematic for time series, where longer time series enable better separation. We mitigate this by leveraging the algorithms from \cite{leonte2024simulation} to simulate fresh training data on-the-fly, rather than repeatedly optimizing the BCE loss on the same potentially separable dataset. 

The second issue is intrinsic to classification 
and persists even under idealized conditions. \cite{bai2021don} prove that logistic regression is overconfident even when the logistic model is well-specified and the sample size far exceeds the parameter count. Thus, while on-the-fly simulation mitigates the finite-sample effect discussed above, it cannot eliminate miscalibration, making post-training calibration essential for 
reliable probability estimates.

\subsubsection{Post-hoc calibration and amortization}
\label{subs:postT_hoc_cal_and_amort}
Recall that $c(\x,\bt) = p(\x,\bt) /\left(p(\x,\bt) + p(\x)p(\bt)\right)$ and that classifier miscalibration directly propagates to the likelihood, posterior distribution, and credible region approximations, as $p(\x \mid \bt) \ \propto \ {p(\bt \mid \x)} \ \propto  \ r(\x,\bt)={\left[\left(c(\x,\bt)\right)^{-1} -1\right]}^{-1}$ for fixed $\x$. To mitigate miscalibration, we apply a post-training monotonic transformation $T \colon [0,1] \to [0,1]$ to the trained classifier $\hat{c}$, yielding $\hat{c}_{\text{cal}} = T \circ \hat{c}$. The transformation is estimated using a separate dataset, allowing it to enforce global properties of the classifier that are difficult to capture from mini-batches during training. 
Based on Theorem $1$ from \cite{Cranmer2015ApproximatingLR}, if $c$ is monotonic with respect to $c^{*}$, in the sense that $c(\x,\bt_1) < c(\x,\bt_2)$ implies $c^{*}(\x,\bt_1) < c^{*}(\x,\bt_2)$, then the optimal $c^{*}$ can be recovered from $c$ by calibration. In practice, we cannot guarantee the monotonicity condition, and we resort to the metrics and checks discussed in the next section to compare performance pre- and post-calibration. 

Another advantage of calibration is amortization over datasets of varying sample size. For example, in the time series context, the neural network used to process the time series observations can be chosen from the class of recurrent neural networks, which accept inputs of variable length. As a result, once a classifier has been trained on time series of fixed length $k$, it can be directly applied to inputs of other lengths $k'$; see Figure~\ref{fig:1} in Section~\ref{subsection:TRE_training_procedure} of the Supplementary Material. However, changing the length acts as a distributional shift that affects the properties of the trained classifier. To account for this, we can fit a calibration mapping on a newly simulated calibration dataset with inputs $\x$ of length $k'$. Intuitively, as $k$ increases, the classifier should be more confident, and the calibration map thus compensates by applying a monotonic correction, which naturally preserves the ranking of likelihoods, and Theorem $1$ from \cite{Cranmer2015ApproximatingLR} guarantees no loss in efficiency. In Section~\ref{section:simulation_study}, we empirically verify this by showing that our TRE yields high-quality approximations even when applied to both shorter and longer time series than those seen during training. The key is that calibration is fast and does not require retraining.

A popular parametric calibration method is Platt scaling \citep{platt1999probabilistic}, which applies the sigmoid transformation $T(s;A,B) = 1/\left(1 + \exp{(-As+B)}\right)$, with $A > 0,B\in\mathbb{R}$. However, Platt scaling is quite rigid: it does not contain the identity, and can sometimes even worsen calibration. For this reason, \cite{kull2017beta} introduce the more flexible family of beta-calibration mappings $T(s;a,b,c) = 1/\left(1 + e^{-c} (1-s)^b  s^{-a} \right)$, with $a,b>0,c\in\mathbb{R}$, which yields the identity for $a = b = 1$, $c = 0$. Beyond parametric approaches, isotonic regression \citep{caruana2004predicting} offers a more flexible, non-parametric alternative. It learns a monotone, stepwise function by solving a quadratic program and can yield superior performance. However, the resulting calibration mapping is piecewise constant, making it unsuitable when differentiability is required, e.g., in gradient-based MCMC. Fortunately, the MCMC-free posterior sampling scheme introduced for TRE in Section~\ref{subsection:mcmc_posterior_sampling} lifts this requirement. In this paper, we test both beta and isotonic calibration for comparison. 
\subsection{Checks on the quality of the approximate likelihood} \label{subsection:posterior_checks}
\subsubsection{Coverage diagnostics}\label{subsection:coverage_diagnostics}
Having trained and calibrated an amortized estimator, we next assess the quality of the approximation $(\x,\bt) \mapsto \hat{r}(\x \mid \bt)$. 
Some approaches include classifier-based tests \citep{lopez2017revisiting,linhart2024c2st} and kernelized Stein divergence tests \citep{kernelized_stein1, chwialkowski2016kernel}, though \cite{lueckmann2021benchmarking} found the above to be sensitive to hyperparameters and even inconsistent with each other. In this paper, we assess the approximation quality by comparing the theoretical and empirical coverage, i.e., we check whether ground-truth parameters fall within prediction regions at the correct rate. 
\begin{definition}\label{def:HPD_main_body} 
Let $\Theta_{\hat{p}(\cdot \, \mid \x )}(1-\alpha)$ be the $1-\alpha$ highest posterior density region (HPD) of the approximate posterior $(\x,\bt) \mapsto \hat{p}(\bt \mid \x)$. The expected coverage at level $1-\alpha$ is 
\begin{equation}
    \mathcal{C}_{1-\alpha} = \ev_{p(\x,\bt)}\left[\mathds{1}\left({\bt}  \in \Theta_{\hat{p}(\cdot \, \mid \x )}(1-\alpha) \right) \right].\label{eq:coverage_def}
\end{equation}
\end{definition} 
We stress that \eqref{eq:coverage_def} can be thought of as both a frequentist expected coverage and a Bayesian credible region, depending on the interpretation of $\ev_{p(\x,\bt)}$ as $\ev_{p(\bt)} \ev_{p(\x \mid \bt)}$ or $\ev_{p(\x)} \ev_{p(\bt \mid \x)}$. If the likelihood approximation $(\x,\bt) \to \hat{p}(\x \mid \bt)$ is faithful, or equivalently if the posterior $(\x,\bt) \to \hat{p}(\bt \mid \x)$ is well calibrated, then $\mathcal{C}_{1-\alpha} = 1-\alpha$ for any $\alpha \in [0,1]$; discrepancies can be visualised by plotting the empirical and theoretical coverage against each other; see Figure~\ref{fig:coverage_comparison} below for illustration. The empirical coverage can be derived as follows. First, generate $N$ samples $(\x_1,\bt_1),\ldots,(\x_N,\bt_N) \stackrel{\text{iid}}{\sim} p(\x,\bt)$. For each $\x_j$, draw $M$ posterior samples  $\vt_{j,1}, \ldots, \vt_{j,M} \sim  \hat{p}(\bt \mid \x_j) = \hat{r}(\x_j \mid \bt) \, p(\bt)$, e.g., using Chebyshev polynomial approximations, and derive the approximate HPD region ${\Theta}^M_{\hat{p}(\cdot \, \mid \x_j )}(1-\alpha) \approx {\Theta}_{\hat{p}(\cdot \, \mid \x_j )}(1-\alpha)$ as follows: sort the values $\hat{p}(\vt_{j,1} \mid \x_j), \ldots,\hat{p}(\vt_{j,M} \mid \x_j )$ in descending order and retain the top $(1-\alpha) M$ of them. The smallest retained posterior density value defines the HPD acceptance threshold. Finally, compare this threshold with $\hat{p}(\bt_j \mid \x_j)$ and calculate the proportion of true parameters that fall within their corresponding HPD regions, i.e.,
\begin{equation}
   \hat{C}_{1-\alpha} = \frac{1}{N}\sum_{j=1}^N\mathds{1}\left( \bt_j \in \Theta^M_{\hat{p}(\cdot \, \mid \x_j )}(1-\alpha)\right) = \frac{1}{N}\sum_{j=1}^N\mathds{1}\left( \hat{p}(\bt_j \mid\x_j) \ge \textrm{threshold}_j \right) .\label{eq:coverage_estimation_empirically}
\end{equation}
We emphasize that the $\Theta^M_{\hat{p}(\cdot \, \mid \x_j)}(1-\alpha)$ region is doubly approximate: it is estimated via samples from the $\hat{p}(\cdot \mid \x)$, which is itself an approximation of the posterior.

\subsubsection{Posterior coverage limitations and novel individual checks}\label{subsection:posterior_coverage_and_novel_checks} 
The coverage check may fail to detect poor approximations in some cases, particularly with non-informative likelihoods. The degenerate case $\hat{p}(\x \mid\vt) \equiv 1$ yields $\hat{p}(\bt \mid \x) = p(\bt)$ and $\mathcal{C}_{1-\alpha} = 1-\alpha$, but can be spotted by inspecting the classifier outputs, which are near $0.5$. More subtle failures may evade immediate detection, such as when classifiers ignore specific components of $\bt$ or when overconfidence and underconfidence across TRE components cancel out, yielding good overall coverage despite misspecified component likelihood ratios. 

To address these limitations, we introduce novel and computationally efficient per-parameter coverage checks. Our approach leverages the TRE framework's access to the conditional densities $\hat{p}(\theta^{i} \mid \x,\bt^{1:i-1})$, which 
enables the construction of HPD regions conditional on both $\x$ and preceding parameters $\bt^{1
:i-1}$, rather than $\x$ alone; once the HPD region is known, coverage is estimated as before, via Equation~\eqref{eq:coverage_estimation_empirically}; see Section \ref{sm:per_parmaeter_checks} in the Supplementary Material for details. 
The key computational advantage lies in our Chebyshev polynomial approach: with typically under $128$ 
Chebyshev knots, we can accurately approximate $\hat{p}(\theta^{i} \mid \x,\bt^{1:i-1})$, generate arbitrarily many samples and compute HPDs near-instantaneously. 
Unlike MCMC iterations, drawing posterior samples 
requires no extra neural network evaluations 
after determining the Chebyshev 
coefficients. Overall, our componentwise approach enables targeted identification of miscalibrated classifiers, as well as model comparison. 
\subsubsection{Unified assessment: calibration, coverage, and metrics}\label{subsectiton:unified_metrics}

As previously discussed, many SBI methodologies, including NRE, produce overly-peaky likelihoods, and equivalently overconfident posteriors \citep{hermans2022trust}. In turn, this is equivalent to the overconfidence of the classifier, as $p(\bt \mid \x) \ \propto  \ r(\x,\bt)={\left[\left(c(\x,\bt)\right)^{-1} -1\right]}^{-1}$. While post-hoc calibration can mitigate miscalibration, it can also sometimes degrade performance, e.g., for NREs distinguishing between samples which can be trivially separated. Hence, it is necessary to carry a case-by-case evaluation and selectively retrain component TRE classifiers when post-hoc calibration alone cannot achieve satisfactory diagnostics. 
To aid 
this analysis, we track metrics during training, before and after calibration. While no single metric fully characterizes posterior quality, monitoring several metrics simultaneously 
helps identify issues and assess convergence. The BCE loss and classification accuracy measure classifier performance. The KL divergence $\mathcal{S} = \ev_{p(\bt,\x)}\left[\log{\hat{r}(\bt \mid \x)}\right]$ 
 enables comparison between SBI estimators, though its logarithmic nature makes it more sensitive to matching density peaks than level curves. Deviations of the balancing metric $\mathcal{B} =\ev_{p(\x, \bt)}\left[\hat{c}(\x, \bt)\right] +\ev_{p(\x)p(\bt)}\left[\hat{c}(\x, \bt)\right]$ from the theoretical value $1$ 
 detects discrepancies of the aforementioned level curves through class imbalances. Finally, the ECE metric from Section~\ref{subsection:potential_reasons} can serve for model comparison.

\section{Simulation study}
\label{section:simulation_study}
Having developed a novel SBI methodology based on TRE and Chebyshev polynomial approximations, we 
assess its performance in a simulation study with trawl process 
time series, which provide a challenging test case for 
point estimation and posterior uncertainty quantification. 
However, note that our methodology applies more broadly to any intractable stochastic process that admits a finite-dimensional parameterization and a fast simulator.
\subsection{Trawl processes and the simulation setting}\label{section:simulation_study_setup}
Trawl processes are a flexible class of continuous-time, non-Markovian stochastic processes constructed as kernel convolutions against infinitely divisible random measures; see Section~\ref{suppl_material:parameterizations} of the Supplementary Material for a formal definition. They generalize Gaussian processes to the infinitely divisible setting and are fully characterized by their marginal distribution and autocorrelation function (ACF). Furthermore, both the marginal and ACF are available in closed form, hence estimation error can be easily computed and interpreted. This makes them particularly suitable for our simulation study. Finally, their likelihood function is generally intractable, and efficient simulation schemes are available \citep{leonte2024simulation}. 

Here, we consider a real-valued trawl process with three-parameter Normal-inverse Gaussian (NIG) marginals
$
    X_t \sim \textrm{NIG}(\mu,\sigma,\beta),
$
where $\mu \in \R$ is the mean, $\sigma > 0$ is the standard deviation and $\beta \in \R$ controls the skewness; see Section~\ref{subsubsection:alternative_specification} of the Supplementary Material for further details. 
For the dependence structure, we use the following ACF:
\begin{equation*}
    \rho(h)
    =
    \exp\left\{
        \eta_{\textrm{acf}}
        \left(
        1-\sqrt{1+\frac{2h}{\gamma_{\textrm{acf}}^2}}
        \right)
    \right\},
    \qquad h\geq 0,
\end{equation*}
with $\gamma_{\textrm{acf}},\eta_{\textrm{acf}}>0$. This yields semi-long memory, i.e., correlation decaying slower than exponential but faster than polynomial. The trawl model is therefore parameterized by
$    \bt =    \left(    \gamma_{\textrm{acf}},
    \eta_{\textrm{acf}},    \mu,\sigma,\beta    \right).
$ 
For each parameter $\bt$, the vector $\x= (X_1,\ldots,X_k)$ denotes an observation of the corresponding trawl process at discrete equidistant time points, and $k$ denotes the length of the observed time series $\x$. Here, we take $k=1500$ during training; see the left panel of Figure~\ref{fig:MLE_comparison_intro} for a simulated trawl process.

\begin{figure}[t]
   \centering
    \includegraphics[width=\textwidth]{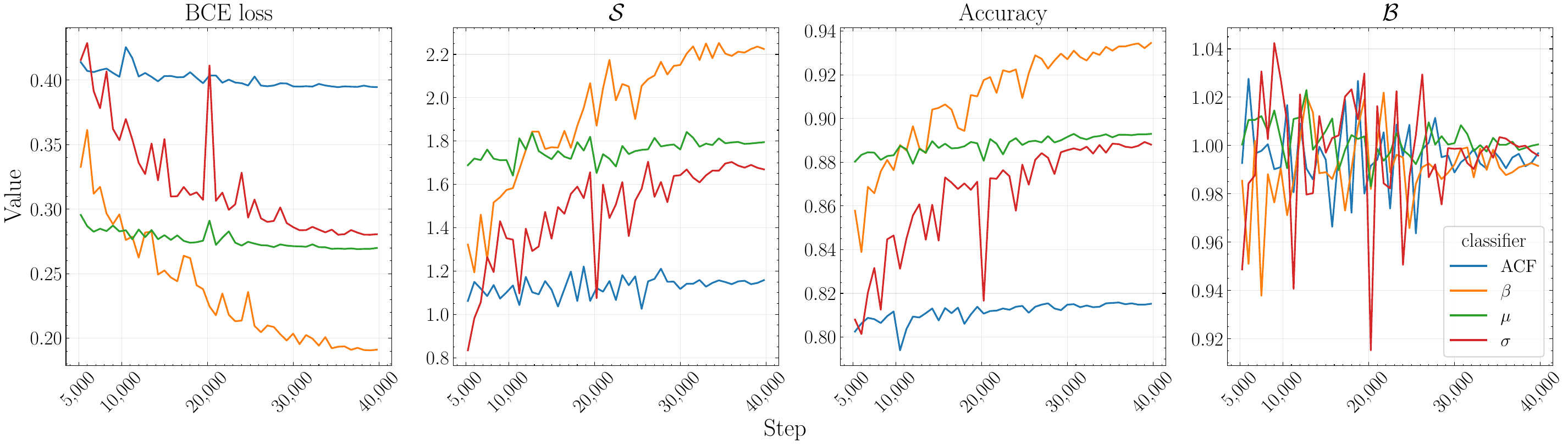}
    \vspace{-0.3cm}
    \caption{BCE, $\mathcal{S}$, classification accuracy (fraction of samples that are correctly classified), and $\mathcal{B}$ metrics (left to right) for the $\textrm{ACF},\, \mu,\, \sigma$ and $\beta$ classifiers (different colored lines), evaluated on a holdout dataset over the last 35000 training iterations. We train the classifiers with trawl process realizations $\x$ of length $1500$. The legend is displayed in the right panel.}
\label{fig:tre_tracking_metrics_during_training}
\end{figure}

We set independent uniform sampling distributions to ensure the amortized model is valid for a wide range of cases: ${\mu \sim \mathcal{U}(-1,1)}, {\sigma \sim \mathcal{U}(0.5,1.5)}$, ${\beta \sim \mathcal{U}(-5,5)}, \gamma_{\text{acf}} \sim \mathcal{U}(10,20)$, and $\eta_{\text{acf}} \sim \mathcal{U}(10,20)$
. The support restrictions on $\mu$ and $\sigma$ are without loss of generality, as we can feed the classifier time series that are centered and scaled with the empirical mean and standard deviation, and back-transform posterior samples afterwards.
As discussed in Section~\ref{section:meth_contr_to_nre}, parameter coordinates can be learnt individually or in blocks. The autocorrelation-function (ACF) parameters are weakly identifiable when considered individually, so we infer them jointly. Learning them separately would introduce an extra classifier with small KL divergence $\mathcal{S}$, thus offering little benefit while increasing the number of classification tasks. The ratio $r(\x,\bt)$ can thus be learnt by training four classifiers:
\begin{align*}
    r(\x,\bt) &= \frac{p(\x,\bt)}{p(\x)p(\bt)}= \frac{p(\x,\gamma_{\textrm{acf}},\eta_{\textrm{acf}},\mu,\sigma,\beta)}{p(\x)\, p(\gamma_{\textrm{acf}}, \eta_{\textrm{acf}}) \, p(\mu) \, p(\sigma) \, p(\beta)} \\
&=  \underbrace{\frac{p(\x,\gamma_{\textrm{acf}},\eta_{\textrm{acf}})}{p(\x)\, p(\gamma_{\textrm{acf}}, \eta_{\textrm{acf}})}}_{\text{ACF classifier}}  
   \cdot \underbrace{\frac{p(\x,\gamma_{\textrm{acf}},\eta_{\textrm{acf}},\mu)}{p(\x,\gamma_{\textrm{acf}},\eta_{\textrm{acf}})\, p(\mu)}}_{\mu \text{ classifier}} \cdot \underbrace{\frac{p(\x,\gamma_{\textrm{acf}},\eta_{\textrm{acf}},\mu,\sigma)}{p(\x,\gamma_{\textrm{acf}},\eta_{\textrm{acf}},\mu)\, p(\sigma)}}_{\sigma \text{ classifier}} \cdot 
\underbrace{\frac{p(\x,\gamma_{\textrm{acf}},\eta_{\textrm{acf}},\mu,\sigma,\beta)}{p(\x,\gamma_{\textrm{acf}},\eta_{\textrm{acf}},\mu,\sigma) \, p(\beta)}}_{\beta \text{ classifier}}\\
&= \underbrace{\frac{p(\gamma_{\textrm{acf}},\eta_{\textrm{acf}} \mid \x )}{p(\gamma_{\textrm{acf}},\eta_{\textrm{acf}})}}_{\text{ACF classifier}}  
   \cdot \underbrace{\frac{p(\mu \mid \x,\gamma_{\textrm{acf}},\eta_{\textrm{acf}})}{p(\mu)}}_{\mu \text{ classifier}} \cdot \underbrace{\frac{p(\sigma \mid \x,\gamma_{\textrm{acf}},\eta_{\textrm{acf}},\mu)}{p(\sigma)}}_{\sigma \text{ classifier}} \cdot
\underbrace{\frac{p(\beta \mid \x,\gamma_{\textrm{acf}},\eta_{\textrm{acf}},\mu,\sigma)}{p(\beta)}}_{\beta \text{ classifier}} .
\end{align*}
For each classifier, the observed time series $\x$ is first processed by a recurrent neural network, specifically a Long Short-Term Memory (LSTM) encoder, which maps the series to a learned, finite-dimensional summary representation. These summary statistics are concatenated with the parameter subset relevant to the corresponding TRE component, i.e., $\bt^{1:i}$, and passed through fully connected layers to obtain the classifier output. Full architectural, computational, and training details are provided in Sections~\ref{subsection:TRE_training_procedure} and~\ref{section:training_details} of the Supplementary Material. We monitor the BCE loss, KL divergence $\mathcal{S}$, classification accuracy (fraction of samples that are correctly classified), and balancing metric $\mathcal{B}$ (see Section~\ref{subsectiton:unified_metrics}) throughout training and display the results in Figure~\ref{fig:tre_tracking_metrics_during_training}. 
All metrics approximately stabilize, suggesting convergence. Gradients are computed with data simulated on-the-fly at each training iteration, whereas the metrics are evaluated on a fixed, holdout dataset. 
Having obtained an approximate mapping $(\x,\bt) \mapsto \hat{p}(\x,\bt)$, we assess its quality when used for neural point estimation and for constructing posterior credible regions. As we show next, our methodology yields substantial improvements over existing approaches across three key dimensions: reduced estimation error, enhanced diagnostic capabilities for assessing the approximation's quality, and amortization across varying time series lengths $k$.
\subsection{Neural point estimation}
\label{subsection:point_estimators}
We begin by numerically deriving maximum likelihood estimates (MLE) from the TRE approximation. For comparison, we consider two benchmarks: MLEs obtained from standard NRE, and point estimates from the generalized method of moments (GMM), which is the traditional inference approach for trawl processes \citep{barndorff2014integer}. For GMM, the marginal parameters are inferred from the first four empirical moments and ACF parameters from the empirical autocorrelation function. Since these methods serve only as benchmarks, we provide their implementation details in Section~\ref{sm:point_estimators} of the Supplementary Material. Figure~\ref{fig:MLE_comparison_intro} illustrates the performance of these methods on a trawl process realization of length $1500$. Note that TRE achieves the closest match to both the true marginal distribution and the ACF structure. 
\begin{figure}[t]
    \centering
    \includegraphics[width=\linewidth]{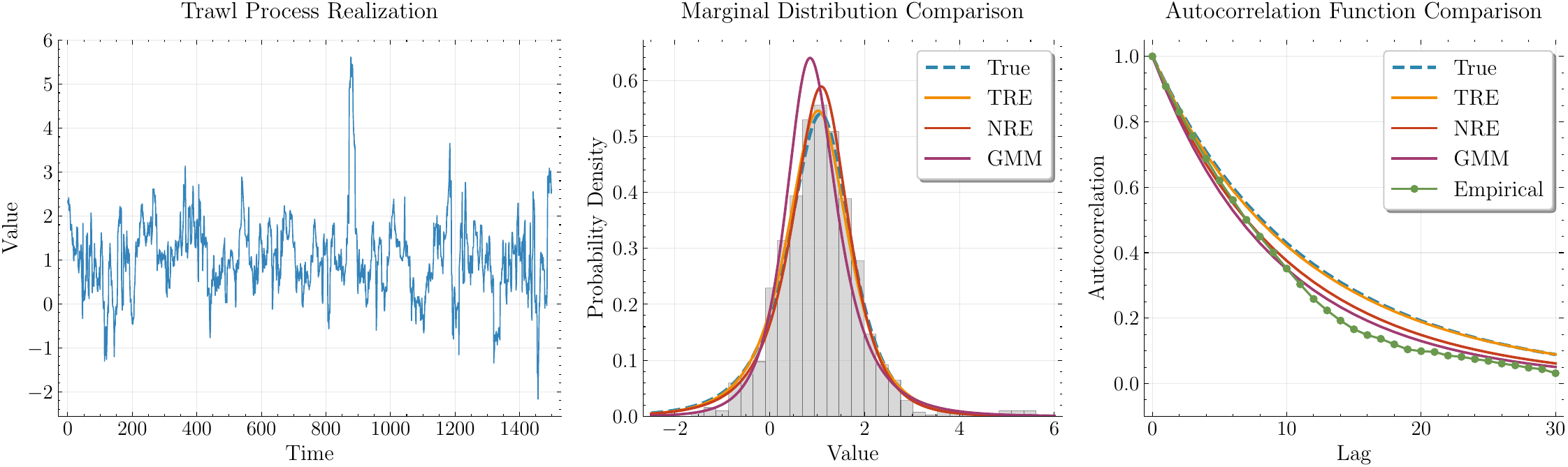}
    \caption{Performance comparison of point estimators given by TRE, NRE and GMM. Left: trawl process realization corresponding to $\bt = (13.36, 15.52,  0.97, 0.98, -0.17)$; middle: true (dashed) and inferred (solid) marginal distributions; right: true (dashed), empirical (solid-dotted), and infered (solid) ACFs. 
    }
    \label{fig:MLE_comparison_intro}
    \vspace{-0.1cm}
\end{figure}
\begin{table}[t]
\centering
\caption{Estimation error comparison across series lengths $k=1000, 1500, 2000$ (top to bottom blocks) for GMM, NRE, TRE, and NBE (top to bottom rows within each block). From left to right, we display mean $L^1$ and $L^2$ distances between the true and inferred ACF functions, MAE and RMSE for the marginal parameters $\mu, \sigma$ and $\beta$, and mean KL divergence between true and inferred marginal distributions. 
}
\scalebox{0.85}{
\begin{tabular}{l@{\hspace{8mm}}l!{\vrule width 1pt}rr!{\vrule width 1pt}rr!{\vrule width 0.3pt}rr!{\vrule width 0.3pt}rr!{\vrule width 0.3pt}r}
\toprule
 &  & \multicolumn{2}{c}{ACF} & \multicolumn{2}{c}{$\mu$} & \multicolumn{2}{c}{$\sigma$} & \multicolumn{2}{c}{$\beta$} & mean KL \\
 &  & mean $L^1$ & mean $L^2$ & MAE & RMSE & MAE & RMSE & MAE & RMSE &  \\
\midrule
 & GMM & 3.491 & 0.618 & 0.224 & 0.335 & 0.249 & 0.323 & 1.368 & 1.918 & 0.431 \\
 & NRE & 1.528 & 0.271 & 0.110 & 0.150 & 0.113 & 0.147 & 0.771 & 1.075 & 0.036 \\
 & TRE & 1.273 & 0.225 & 0.097 & 0.133 & 0.090 & 0.119 & 0.637 & 0.870 & 0.025 \\
\multirow{-4}{*}{1000}  & NBE & 1.197 & 0.211 & 0.101 & 0.135 & 0.089 & 0.115 & 0.529 & 0.694 & 0.040 \\
\midrule
 & GMM & 3.118 & 0.552 & 0.193 & 0.297 & 0.225 & 0.298 & 1.289 & 1.852 & 0.370 \\
 & NRE & 1.180 & 0.210 & 0.087 & 0.117 & 0.096 & 0.128 & 0.593 & 0.827 & 0.024 \\
 & TRE & 1.075 & 0.190 & 0.082 & 0.112 & 0.078 & 0.102 & 0.558 & 0.770 & 0.017 \\
 \multirow{-4}{*}{1500} & NBE & 1.006 & 0.178 & 0.087 & 0.114 & 0.077 & 0.098 & 0.458 & 0.605 & 0.029 \\
\midrule
 & GMM & 2.825 & 0.500 & 0.175 & 0.274 & 0.205 & 0.274 & 1.249 & 1.817 & 0.326 \\
 & NRE & 1.181 & 0.209 & 0.085 & 0.114 & 0.090 & 0.118 & 0.630 & 0.902 & 0.021 \\
 & TRE & 0.937 & 0.166 & 0.074 & 0.100 & 0.070 & 0.091 & 0.507 & 0.706 & 0.014 \\
\multirow{-4}{*}{2000} & NBE & 0.899 & 0.159 & 0.079 & 0.103 & 0.072 & 0.091 & 0.417 & 0.552 & 0.025 \\
\bottomrule
\end{tabular}
}
\label{table:mle}
\end{table}

To confirm the superiority of TRE, we conduct a simulation study using $10^4$ samples $(\x,\bt)$ drawn from the model. For each realization, we determine TRE-, NRE- and GMM-based point estimates via the BFGS optimization routine. While all methods require an initial value, for TRE a suitable starting point could be obtained efficiently by drawing a small number of posterior samples and selecting the one with highest posterior density. However, to make a fair comparison between the different estimation methods and eliminate sensitivity to initialization, we always start BFGS at the true parameter $\bt$. The estimation error is summarized in Table~\ref{table:mle}. We treat marginal and ACF parameters separately.  For $(\mu,\sigma,\beta)$, we report the mean-absolute error (MAE), root mean-squared error (RMSE), and further provide the KL divergence between the true and inferred marginal distributions. For $\left(\gamma_{\textrm{acf}},\eta_{\textrm{acf}}\right)$, we directly compare the resulting ACFs using the $L^1$ and 
$L^2$ distances rather than comparing individual parameters, as vastly different parameter values can sometimes yield nearly identical ACFs, i.e., these parameters are weakly identifiable. While both TRE and NRE are trained on time series of length $k=1500$, the LSTM encoder enables inference with time series of arbitrary length, so we also test it with time series of length $k=1000$ and $k=2000$. 
Accross all metrics and lengths $k$, TRE substantially outperforms NRE and GMM. A natural question is how TRE compares to neural Bayes estimators (NBE), customized to target point summaries of the posterior distribution \citep{matt_american_statistician}. Training NBEs for trawl processes is tricky, as optimizing for the parameter-wise MAE or MSE fails due to the ACF weak-identifiability. We therefore train NBEs with multiple losses; see Section~\ref{sm:point_estimators} of the Supplementary Material for details. Table~\ref{table:mle} shows the results for the NBE trained with $L^2$ ACF distance and MSE loss for marginal parameters. Even though NBEs have the advantage of directly optimizing the evaluation metric during training, we see that TRE achieves comparable or superior performance while providing the full posterior rather than just point estimates.
\subsection{Calibration, coverage, and posterior checks}
\label{section:calibration_coverage_posterior_checks}
Although TRE point estimators remain accurate when the TRE model trained on time series of length $k=1500$ is applied to time series of other lengths, the encoder's hidden state may encode length-specific features through its weights and biases. As we show below, the level sets of the likelihood approximation indeed become distorted as the input length varies. To address this problem and amortize the estimators across different input lengths, we apply beta-calibration to the NRE and TRE outputs based on multiple datasets $(\x, \bt)$. Specifically, we take the approximate NRE and TRE models likelihood-ratios with $k = 1500$ and calibrate them using datasets where the trawl process realizations $\x$ have lengths $k =1000$, $1500$ and $2000$. Importantly, while calibration modifies the geometry of the likelihood approximation, the estimation-errors from Table~\ref{table:mle} are unchanged (results not shown for brevity). We evaluate the accuracy of the resulting likelihood approximation both before and after calibration using the coverage diagnostics from Section~\ref{subsection:coverage_diagnostics} and metrics from Section~\ref{subsectiton:unified_metrics}. 
\begin{figure}[t]
    \centering
\includegraphics[width=\linewidth]{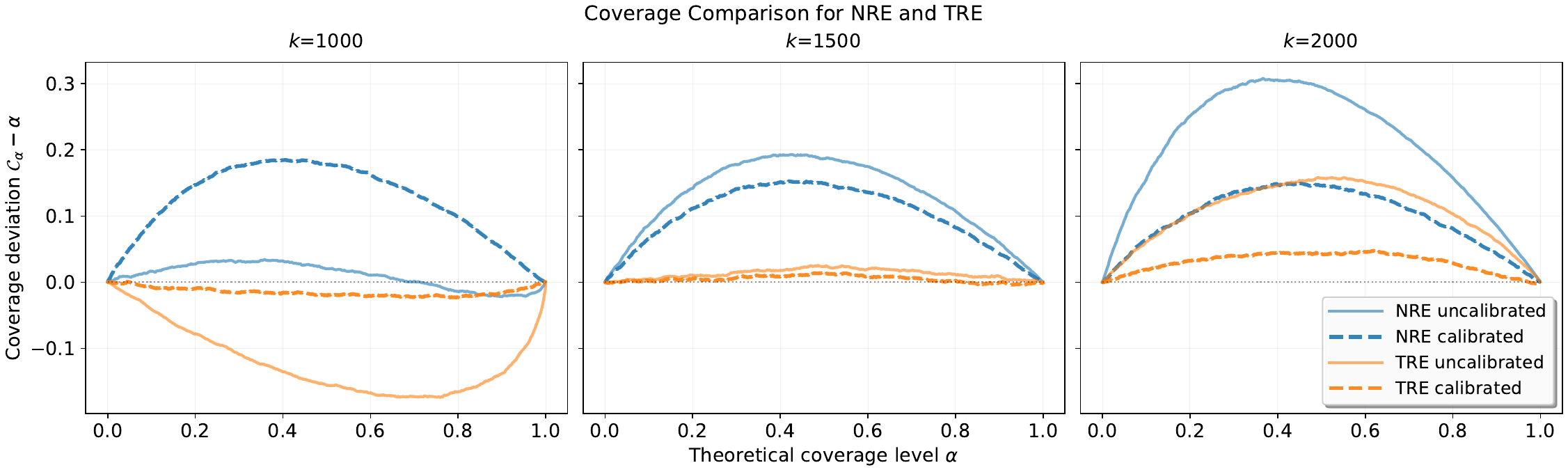}\\[1em]  
    \includegraphics[width=\linewidth]{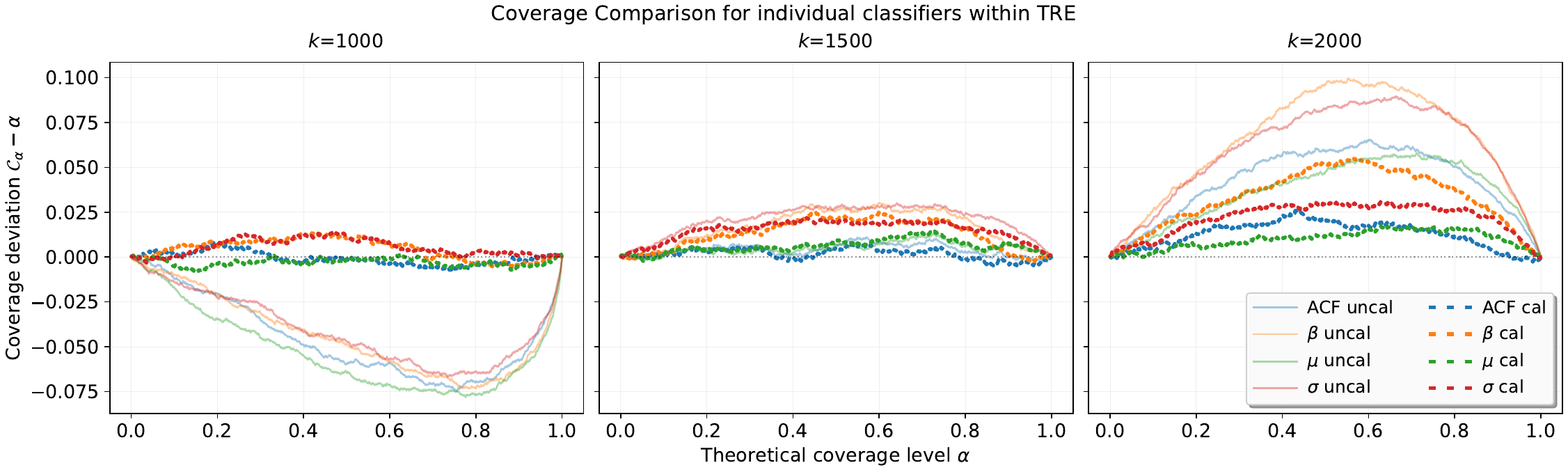}
    \caption{Comparison of the coverage deviation $\mathcal{C}_{\alpha} - \alpha$ before (solid curves) and after (dashed curves) beta-calibration. Positive values indicate underconfidence, while negative ones indicate overconfidence. Top row: NRE (blue) and TRE (orange). Bottom row: component NREs within TRE. Beta-calibration yields near-perfect coverage for 
    TRE, consistently improving TRE performance at all lengths $k=1000$, $1500$, $2000$ (left to right). 
}
    \label{fig:coverage_comparison}
\end{figure}
Figure~\ref{fig:coverage_comparison} displays the deviations $\mathcal{C}_{\alpha} - \alpha$, representing departures from perfect calibration across the trained classifiers.  Starting with the bottom row of the figure, we show results for the ACF, $\mu$, $\sigma$ and $\beta$ classifiers used within the TRE. These classifiers vary in their degree of miscalibration and thus require different adjustments. As discussed in Section~\ref{subsection:NRE_learning_likelihood_through_classification}, the higher the $\mathcal{S}$ value, the harder it is to learn level curves. Thus, the ACF classifier requires the least calibration and also has the lowest $\mathcal{S}$ value, see Figure~\ref{fig:tre_tracking_metrics_during_training}. Turning to the top row of the figure, we compare NRE and TRE. In the case of TRE, calibration is applied individually to each component classifier rather than to the combined estimator as a whole. Individual calibration addresses the distinct miscalibration levels across components, whereas calibrating the combined estimator applies a correction that fails to account for component-specific differences. Further, as the combined classifier's outputs are very close to $0$ or $1$, calibration can become unreliable: see NRE for $k=1000$, where coverage degrades when calibration is applied. Overall, the calibrated TRE has near perfect coverage, and outperforms both uncalibrated TRE and (uncalibrated or calibrated) NRE. 
\begin{table}[t!]
\centering
\caption{Comparison of NRE, TRE and (ACF, $\beta$, $\mu$, $\sigma$) TRE components (left to right) across different series lengths $k=1000$, $1500$, $2000$ (top to bottom blocks) based on the BCE, $\mathcal{S}$, $\mathcal{B}$, $W$ and ECE metrics (top to bottom rows within each block). Values are shown before (uncal) and after (cal) beta-calibration. For a given estimator, lower BCE, $W$, and ECE and higher $\mathcal{S}$ indicate better performance, while $\mathcal{B}$ should be close to its theoretical value $1$. 
}
\scalebox{0.85}[0.85]{%
\begin{tabular}{l@{\hspace{8mm}}l!{\vrule width 0.8pt}rr!{\vrule width 0.3pt}rr!{\vrule width 1pt}rr!{\vrule width 0.3pt}rr!{\vrule width 0.3pt}rr!{\vrule width 0.3pt}rr}
\toprule
 &  & \multicolumn{2}{c}{NRE} & \multicolumn{2}{c}{TRE} & \multicolumn{2}{c}{ACF} & \multicolumn{2}{c}{$\beta$} & \multicolumn{2}{c}{$\mu$} & \multicolumn{2}{c}{$\sigma$} \\
 &  & uncal & cal & uncal & cal & uncal & cal & uncal & cal & uncal & cal & uncal & cal \\
\midrule
& BCE & 0.039 & 0.036 & 0.036 & 0.025 & 0.441 & 0.430 & 0.230 & 0.223 & 0.308 & 0.299 & 0.325 & 0.317 \\
& $\mathcal{S}$ & 5.455 & 5.238 & 6.038 & 6.177 & 0.967 & 0.995 & 2.026 & 2.064 & 1.576 & 1.615 & 1.468 & 1.504 \\
\multirow{-1}{*}{1000}  & $\mathcal{B}$& 0.987 & 1.000 & 0.978 & 0.999 & 0.960 & 1.000 & 0.968 & 0.999 & 0.961 & 1.000 & 0.964 & 1.000 \\
 & $W$ & 0.018 & 0.124 & 0.119 & 0.015 & 0.044 & 0.003 & 0.043 & 0.006 & 0.052 & 0.003 & 0.040 & 0.006 \\
 & ECE & --- & --- & --- & ---  & 0.068 & 0.015 & 0.114 & 0.016 & 0.095 & 0.016 & 0.083 & 0.013 \\
\midrule
 & BCE & 0.023 & 0.022 & 0.015 & 0.015 & 0.388 & 0.388 & 0.199 & 0.199 & 0.268 & 0.268 & 0.285 & 0.285 \\
 & $\mathcal{S}$ & 6.034 & 5.914 & 6.889 & 6.949 & 1.178 & 1.186 & 2.237 & 2.267 & 1.789 & 1.805 & 1.685 & 1.690 \\
\multirow{-1}{*}{1500} & $\mathcal{B}$ & 1.005 & 1.000 & 1.000 & 1.000 & 1.000 & 1.000 & 1.000 & 1.001 & 0.998 & 1.001 & 1.005 & 1.002 \\
 & $W$ & 0.129 & 0.100 & 0.013 & 0.006 & 0.004 & 0.003 & 0.017 & 0.013 & 0.005 & 0.006 & 0.021 & 0.015 \\
 & ECE & --- & --- & --- & --- & 0.003 & 0.004 & 0.013 & 0.008 & 0.006 & 0.006 & 0.012 & 0.009 \\
\midrule
 & BCE & 0.020 & 0.016 & 0.012 & 0.010 & 0.368 & 0.364 & 0.189 & 0.186 & 0.253 & 0.250 & 0.268 & 0.264 \\
 & $\mathcal{S}$ & 6.289 & 6.374 & 7.298 & 7.442 & 1.276 & 1.309 & 2.338 & 2.389 & 1.895 & 1.926 & 1.788 & 1.818 \\
\multirow{-1}{*}{2000} & $\mathcal{B}$ & 1.009 & 1.000 & 1.004 & 1.000 & 1.021 & 1.001 & 1.015 & 1.001 & 1.016 & 1.001 & 1.022 & 1.001 \\
 & $W$ & 0.205 & 0.097 & 0.106 & 0.030 & 0.042 & 0.013 & 0.064 & 0.033 & 0.036 & 0.010 & 0.059 & 0.022 \\
 & ECE & --- & --- & --- & --- & 0.054 & 0.006 & 0.100 & 0.011 & 0.063 & 0.008 & 0.080 & 0.010 \\
\bottomrule
\end{tabular}
}
\label{table:calibration_metrics}
\end{table}

Next, Table~\ref{table:calibration_metrics} displays the BCE, $\mathcal{S}$, $\mathcal{B}$ and expected calibration error (ECE) metrics. We also report $W = \int_0^1 \left|\mathcal{C}_\alpha - \alpha \right| \mathrm{d}\alpha $, which measures the overall absolute deviation between the empirical and theoretical coverage. This metric corresponds to the Wasserstein distance between the distribution of posterior ranks and the uniform distribution, and can be interpreted as a rank check; see Section~\ref{SM:ranks} of the Supplementary Material for further details. The calibrated TRE performs best across all metrics, with beta-calibration consistently improving individual classifier performance. An interesting case is the trained NRE applied to time series of length $k=1000$: the inherent underconfidence observed at $k=1500$ counterbalances, on average, the overconfidence gained from inputting shorter time series. However, the BCE and $\mathcal{S}$ metrics reveal that despite the reasonable overall coverage, the classifier's quality is substantially worse than the corresponding TRE with $k=1000$. 
\begin{remark}\label{remark:posterior_integrals}
In this section, we employed beta-calibration to make a fair comparison between TRE and NRE. While isotonic regression is more flexible, it prevents gradient-based MCMC, thereby efficient posterior sampling and inference for the NRE. By contrast, for TRE, the advocated Chebyshev polynomial approach is compatible with isotonic regression, and Section~\ref{SM:extended_beta_vs_iso} of the Supplementary Material shows that it 
can outperform beta-calibration. 
Methodologically, note that the approximate densities $\theta^{i} \to \hat{p}(\theta^{i} \mid \x,\bt^{1:i-1})$ may not integrate exactly to $1$. Sequential sampling in the dimensions of $\bt$ may thus yield different results to those obtained by combining the approximate ratios within MCMC, as the NRE components' weights differ. We do not observe significant deviations 
 and defer this to future research. 
\end{remark}

\section{Application}
\label{section:application}
\begin{figure}[t!]
    \includegraphics[]{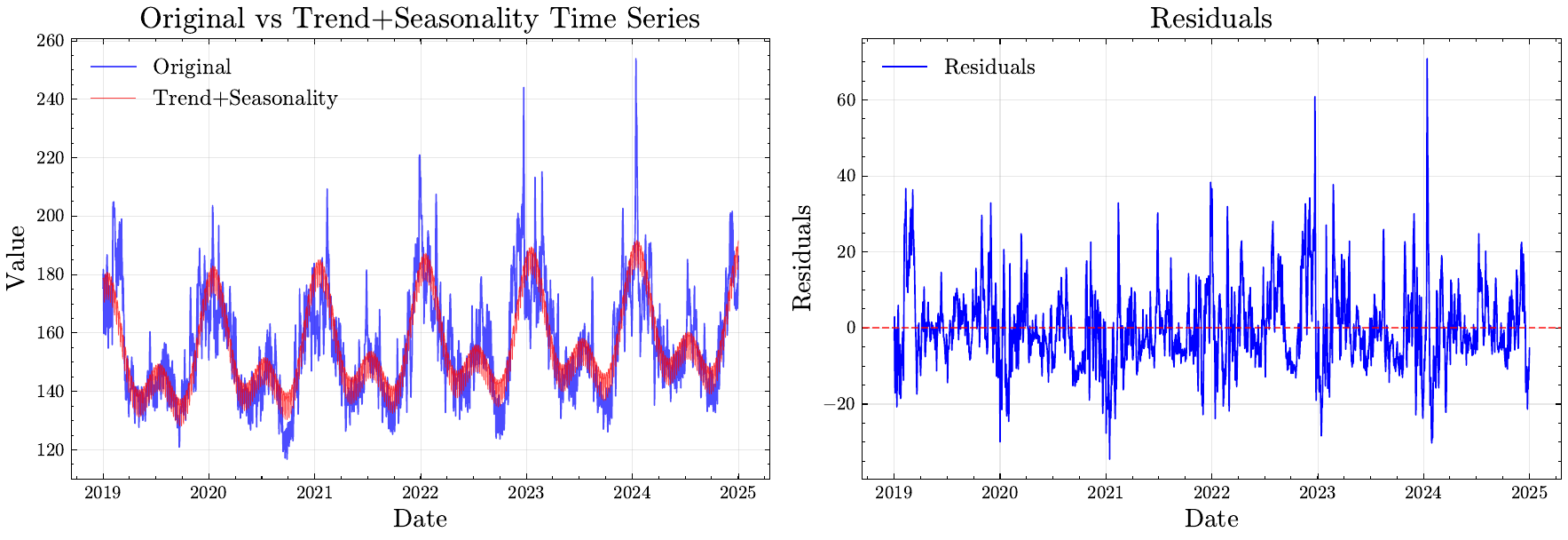}
    \caption{Electricity demand in gigawatt-hours as reported by Bonneville Power Administration Transmission (BPAT). Left: original demand time series together with estimated trend and seasonality; right: residuals.}
\label{figure:application_original_vs_trend_and_seasonality}
\end{figure}

We apply our methodology to daily energy demand data from January 1, 2019, to December 31, 2024, across nine U.S. regions, reported by respondents with acronyms AZPS, BPAT, CISO, DUK, ERCO, FPL, MISO, NYIS, and PJM. To account for non-stationary behaviour, each time series is deseasonalised by ordinary least squares regression using a linear time trend, two annual Fourier harmonics with period $365.25$ days, and day-of-week indicator variables. 
The \texttt{Python} code used to download the datasets and perform the analysis is provided alongside the paper. 
Figure~\ref{figure:application_original_vs_trend_and_seasonality} shows the decomposition of the time series from Bonneville Power Administration Transmission (BPAT) into two parts: a non-stationary component, which captures the trend and seasonal fluctuations, and a stationary component, represented by the residuals after removing trend, seasonality and day-of-week effects.

\begin{figure}[t!]
    \centering
    \includegraphics[]{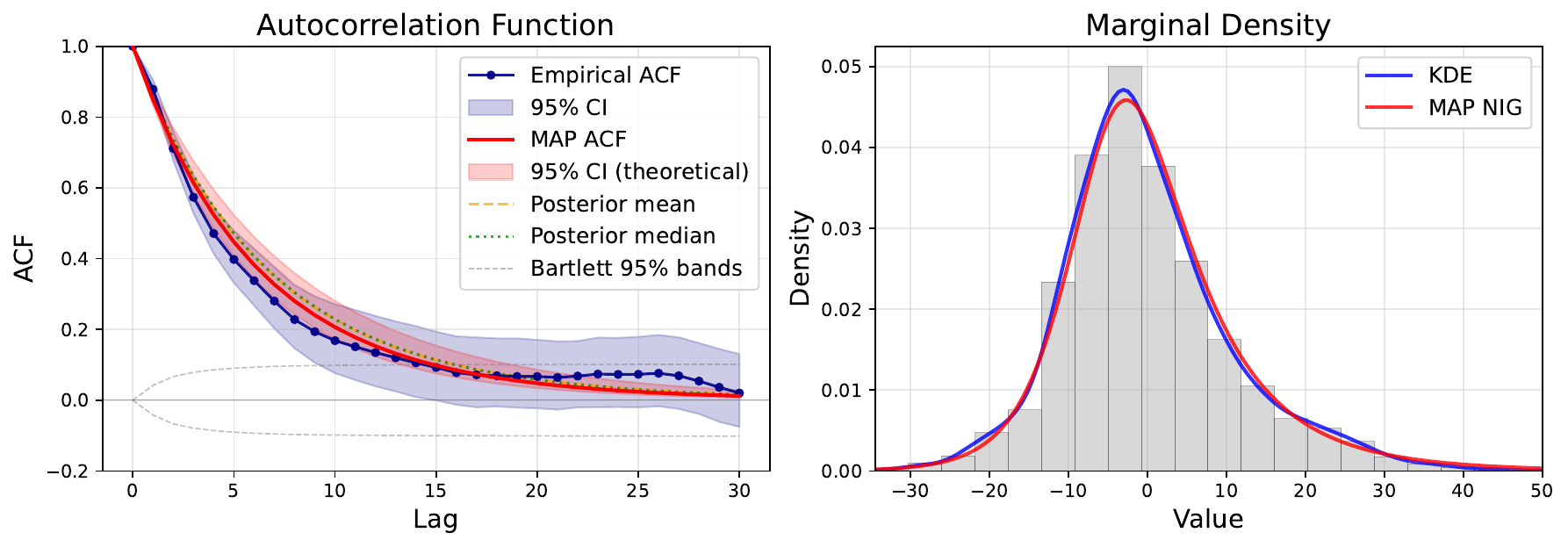}
    \caption{Left: Histogram, kernel density estimate (KDE), and the NIG marginal distribution corresponding to the MAP-fitted parameters of the trawl process. Right: Empirical autocorrelation function (ACF) with $95\%$ bootstrapped credible intervals (CI), the MAP-inferred ACF corresponding to the MAP-fitted parameters with its $95\%$ theoretical CI, and the posterior mean and median ACFs. The posterior mean and median ACFs are computed by taking the lag-wise mean and median of the ACFs evaluated at parameter draws from the posterior distribution of the trawl process parameters. Bartlett’s CI are 
    shown for reference.}
    \label{fig:application_fitted_trawl}
\end{figure}

We use the TRE classifiers trained in the simulation study, without retraining, to infer the parameters of the trawl process from the residuals of all nine time series, and present the results for BPAT in Figure~\ref{fig:application_fitted_trawl}. We also summarize results for all time series in Table~\ref{table:application}. Among the findings, note that the skew parameter $\beta$ is different from $0$ at the $95\%$ level in six out of the nine datasets. The amortized TRE posterior sampling based on Chebyshev polynomial approximations is highly efficient: generating $10^{3}$ independent samples takes under one second on a single Intel i9-14900K (3.20 GHz) CPU core, with further acceleration possible on a GPU, enabling real-time inference. Another advantage is that we avoid costly gradient-descent runs with multiple initializations to accurately determine the MAP, since the maximum over the posterior samples already provides a reliable starting point. 
\begin{table}[!ht]
\centering
\caption{MAP estimates of the trawl process parameters (columns): $\mu, \sigma$, $\beta$, as well as the lag-$1$ correlation and the effective correlation range, $\tau_{\text{eff}}$, defined as the lag at which the MAP-inferred ACF falls below $0.05$, across nine energy-demand time series (rows). Numbers in brackets are $95\%$ credible intervals.}
\resizebox{0.915\textwidth}{!}{%
\begin{tabular}{lcccccc}
\toprule
Dataset & $\mu$ & $\sigma$ & $\beta$ & ACF(1) & $\tau_{\text{eff}}$ \\
\midrule
AZPS & -0.12 (-1.47, 1.49) & 8.51 (7.80, 9.46) & 0.06 (-0.13, 0.27) & 0.88 (0.86, 0.90) & 25 \\
ERCO & 7.65 (-9.18, 24.63) & 115.28 (107.09, 130.42) & 0.39 (0.15, 0.81) & 0.84 (0.83, 0.86) & 18 \\
CISO & -0.59 (-6.17, 7.67) & 42.93 (39.88, 47.72) & 0.09 (-0.08, 0.31) & 0.86 (0.84, 0.88) & 21 \\
NYIS & -0.17 (-3.69, 7.07) & 35.26 (33.82, 39.76) & 0.24 (0.04, 0.50) & 0.84 (0.82, 0.85) & 17 \\
PJM & 0.99 (-16.29, 38.16) & 185.37 (177.83, 212.51) & 0.34 (0.14, 0.64) & 0.84 (0.82, 0.86) & 17 \\
MISO & 0.94 (-14.10, 22.08) & 115.77 (111.19, 132.37) & 0.25 (0.09, 0.49) & 0.85 (0.83, 0.87) & 18 \\
FPL & -2.79 (-6.97, 3.84) & 35.96 (34.45, 40.36) & -0.19 (-0.45, 0.05) & 0.84 (0.82, 0.86) & 17 \\
DUK & 0.37 (-2.78, 6.31) & 32.15 (30.46, 35.91) & 0.42 (0.18, 0.79) & 0.83 (0.82, 0.84) & 17 \\
BPAT & 0.61 (-1.13, 2.52) & 11.77 (10.45, 13.32) & 0.54 (0.26, 0.86) & 0.86 (0.83, 0.88) & 20 \\
\bottomrule
\end{tabular}%
}
\label{table:application}
\end{table}

Although the fitted marginal distribution and ACF for the BPAT dataset agree well with their empirical counterparts, see Figure~\ref{fig:application_fitted_trawl}, it is not obvious if the preprocessing steps fully removed the seasonal trends and heteroskedasticity, and if the residuals truly follow the law of a stationary trawl process. Developing goodness-of-fit tests that build upon our SBI framework would allow for modelling assumptions to be checked directly.

\section{Conclusion}
\label{section:conclusion}
In this paper, we advance the existing neural ratio estimation methodology for amortized simulation-based inference (SBI) along four directions. First, we propose telescoping ratio estimation (TRE) in the SBI setup and decompose the global classification task into multiple subtasks, thereby improving training efficiency and mitigating the ``curse of parameter space dimensionality''. Second, we propose a novel, MCMC-free sequential sampling approach that is GPU-compatible and that leverages Chebyshev polynomial approximations for efficient inverse sampling. Third, we develop novel, granular and nearly instantaneous diagnostic checks on the quality of the learnt likelihood ratio. Beyond faster inference, it also makes large-scale model comparison practically feasible. Fourth, we show that post-training calibration can be integrated into TRE to both improve the learnt ratio and further extend amortization, for instance, across different lengths of the input time series. Through extensive simulations and an application to energy demand data based on non-Gaussian trawl process models, we demonstrated significant improvements over existing methods and performance on par with neural Bayes estimators.

Our work also opens further research avenues. From a modelling perspective, the methodology can be directly extended both to the non-stationary setting and to the spatial or spatio-temporal setting. Periodic trawl processes \citep{periodic_trawls} provide a natural way to incorporate seasonal effects directly into the model. Moreover, by indexing the sets $A_t(\mathbf{z})$ in both time $t$ and space $\mathbf{z} \in \R^d$, the framework easily extends to spatio-temporal models of the form $(t,\mathbf{z}) \to L(A_t(\mathbf{z}))$. These two directions can also easily be combined, giving rise to non-stationary, spatio-temporal models that can capture both seasonal variation and spatial dependence and that still admit fast simulators. From an inferential perspective, amortization could encompass not only varying series lengths but also irregular spacing and partial observations while preserving the reliability of the likelihood estimates. Finally, future work could also benchmark our approach against other SBI methods to better understand its comparative advantages.

\par
\textbf{Disclosure statement.} The authors have no conflict of interests to report.

\textbf{Acknowledgments.} The authors would like to thank Hugo Chu and Maja Schneider for helpful discussions and comments on the manuscript. Claude 3 (Opus, Sonnet, Haiku) and GPT-4o and 5.6 were used in this work, mainly for proofreading, rephrasing written text and improving phrasing, and refactoring Python computer code. This publication is based upon work supported by King Abdullah University of Science and Technology (KAUST) under Award No. RFS-OFP2023-5550 and by EPSRC Centre for Doctoral Training in
Mathematics of Random Systems: Analysis, Modelling and Simulation (EP/S023925/1).

\textbf{Data availability statement.} 
 The dataset used in the the application, i.e., Section~\ref{section:application} is available from \url{https://www.eia.gov/}. We also provide code to download these data, and reproduce the results from this paper at \url{https://github.com/danleonte/Simulation-based-inference-via-telescoping-ratio-estimation-for-trawl-processes-paper}.

{\setstretch{0.89}%
 \setlength{\bibsep}{4pt plus 1pt}
 \setlength{\bibhang}{0.5em}
 \bibliography{bibliography}}

\clearpage
\newpage

\phantomsection\label{supplementary-material}
\bigskip

\renewcommand{\thesection}{S\arabic{section}}
\renewcommand{\thetable}{S\arabic{table}}
\renewcommand{\thefigure}{S\arabic{figure}}
\renewcommand{\theequation}{S\arabic{equation}}

\setcounter{theorem}{0}
\setcounter{section}{0}
\setcounter{figure}{0}
\setcounter{table}{0}
\setcounter{equation}{0}

\setcounter{page}{1}
\renewcommand{\thepage}{S\arabic{page}}

\setcounter{algorithm}{0}
\renewcommand{\thealgorithm}{S\arabic{algorithm}}

\begin{center}
    {\Large\bf Supplementary Material for ``Simulation-based inference via telescoping ratio estimation for trawl processes''}
\end{center}

\vspace{10pt}
\begin{center}
    {\large\bf Dan Leonte\footnote{Computer, Electrical and Mathematical Sciences and Engineering (CEMSE) Division, King Abdullah University of Science and Technology (KAUST), Thuwal, Saudi Arabia; emails: dan.leonte@kaust.edu.sa, raphael.huser@kaust.edu.sa}, Rapha\"el Huser\footnotemark[1], and Almut E. D. Veraart\footnote{Department of Mathematics, Imperial College London, London, United Kingdom; email: a.veraart@imperial.ac.uk}}
\end{center}
\vspace{20pt}

The Supplementary Material is organized as follows.

\begin{itemize}
\item Section~\ref{suppl_material:parameterizations} provides the background on L\'evy bases and trawl processes in~\ref{subsection:levy_bases} and~\ref{subsection:trawl_background}, respectively, together with a list of the probability distribution parameterizations used throughout the main text in~\ref{subsubsection:parameterizations} and~\ref{subsubsection:alternative_specification}.

\item Section~\ref{appendix:subsection_tre} provides the pseudo-code for the ratio estimation procedures from Section~\ref{sec-meth}, 
and a discussion on neural network architectures for TRE in Section~\ref{subsection:TRE_training_procedure} (with further training details are provided in~\ref{section:training_details}). The proof for Theorem~\ref{prop:kl_decomposition} is given in~\ref{subsection:proof_of_proposition}.

\item Section~\ref{suppl:subsection_chebyshev} introduces TRE posterior sampling based on Chebyshev polynomials approximations. We first review the approximation theory in the orthonormal basis of Chebyshev polynomials in Section~\ref{SM:cheb_background}, then show how this can be applied to sample from univariate and bivariate densities known only up to a normalizing constant, effectively replacing MCMC methods in Section~\ref{SM:sampling_from_1d_and_2d}. This sampling approach is then used for coverage checks in Sections~\ref{sm:per_parmaeter_checks} and \ref{SM:seq_p_sam}.

\item Section~\ref{SM:calibration_and_coverage} gives further details on the approximation of the expected calibration error (ECE) and rank checks.

\item Section~\ref{SM:extended_sim_study} presents the extended simulation study. In Section~\ref{sm:point_estimators}, we expand the point estimator comparison from Section~\ref{subsection:point_estimators}, detailing the methodologies for generalized method of moments (GMM), neural ratio estimation (NRE), and neural Bayes estimators (NBE). We show that TRE achieves optimal or near-optimal performance across all the evaluation metrics, whereas the NBEs performance depends strongly on both the training objective and evaluation metric. In Section~\ref{SM:extended_beta_vs_iso}, we extend the calibration analysis from Section~\ref{section:calibration_coverage_posterior_checks}. Specifically, in Section~\ref{section:calibration_coverage_posterior_checks} we compared NRE and TRE with the corresponding beta-calibrated versions. Nevertheless, the sampling techniques based on Chebyshev polynomial approximations enable isotonic regression calibration for TRE without detering posterior sampling, and we compare beta-calibration with isotonic regression for TRE in Section~\ref{SM:extended_beta_vs_iso}. Finally, Section~\ref{section:training_details} presents model architectures.
\end{itemize}
The Python code used to download data, carry the analysis and reproduce the figures and tables in this paper is available at 
\url{https://github.com/danleonte/Simulation-based-inference-via-telescoping-ratio-estimation-for-trawl-processes-paper}.
\newpage
\section{Background on L\'evy on trawl processes and list of probability distribution parameterizations}
\label{suppl_material:parameterizations}
Trawl processes are stochastic processes constructed by integrating a deterministic kernel against a random measure. The random measure, called a L\'evy basis, is the source of randomness, while the kernel controls how past contributions are aggregated over time. We begin by introducing Lévy bases and subsequently define trawl processes.
\subsection{L\'evy bases} \label{subsection:levy_bases}
Let $\mathcal{B}_{\text{Leb}}(\R^d)$ be the restriction of the Borel $\sigma$-algebra on $\R^d$ to sets of bounded Lebesgue measure. 
\begin{definition}
\label{def:levy_basis}
A L\'evy basis $L$ is a collection of infinitely-divisible, real-valued random variables $\left\{L(A): A \in \mathcal{B}_\text{Leb}(\R^d)\right\}$ such that for any disjoint sets $A,B \in \mathcal{B}_\text{Leb}(\R^d),$ the random variables $L(A)$ and $L(B)$ are independent and $L\left(A \cup B\right)= L(A) + L(B)$ almost surely.
\end{definition}
We focus on stationary L\'evy bases, defined by the property that for any $\uu \in \R^d$ and $A \subset \R^d$, $L(A+\uu)$ and $L(A)$ are equal in law, where $A+\uu$ is the set addition.  \cite{ambit_book} shows that stationary L\'evy bases are homogeneous, meaning the law of $L(A)$ depends on $A$ only through its area $\textrm{Leb}(A)$. Further, \cite{pedersen2003levy} extends the L\'evy-Ito decomposition from L\'evy processes to bases, showing that $L$ can be written as the sum of independent Gaussian and jump L\'evy bases, i.e., $L=L_g+L_j$ with $L_g(A) \sim \mathcal{N}\left(\mu \, \textrm{Leb}(A), \sigma^2 \, \textrm{Leb}(A)\right)$ for some drift $\mu$, variance $\sigma^2 $ and compensated jump process $L_j(A)$. 

For convenience, we call a random variable $L'$ a L\'evy seed for $L$ if $L'$ has the same law as $L\left([0,1]^d\right)$. 
More generally, any $L(A_1)$ is again a L\'evy seed for any set $A_1 \in \mathcal{B}_\text{Leb}(\mathbb{R}^d)$ with $\mathrm{Leb}(A_1) = 1$. The notion of L\'evy seeds allows us to state the distributional properties of $L$ without reference to a specific set and provides a flexible class of marginal distributions, supported on the integers, 
reals 
or positive reals. 
A popular example encompassing many families of L\'evy seeds is the class of generalized hyperbolic distributions \citep{Borak2011}, denoted $\textrm{GH}(\lambda,\alpha,\beta,\delta,\mu)$, which has semi-heavy tails and density given by
\begin{equation*}
x \mapsto \frac{\left(\gamma/\delta\right)^{\lambda}}{\sqrt{2 \pi} K_{\lambda}(\delta \gamma)}\frac{ K_{\lambda-1/2}\left(\alpha   \sqrt{\delta^2 + (x - \mu)^2}\right)}{\left(\sqrt{\delta^2 + (x - \mu)^2}/\alpha \right)^{1/2 - \lambda}}                   e^{\beta  (x - \mu)}, \text{ for } \alpha, \delta \in \R^{+} \text{ and } \beta, \mu, \lambda  \in \R, 
\end{equation*}
and where $\gamma \defeq \sqrt{\alpha^2 - \beta^2}$ and $K_{\lambda}$ is the modified Bessel function of the second kind of order $\lambda$. This family includes many standard distributions that are supported on the reals as special or limiting cases and offers great flexibility. From a statistical modelling point of view, it is often advantageous to have the marginal law of $L(A)$ in closed form. Many common infinitely divisible distributions (e.g., Poisson, Gaussian, and Gamma) are closed under convolutions, meaning that $L'$ and $L(A)$ are from the same named family: for example, if $L'\sim \textrm{Poisson}(\nu)$, then $L(A) \sim \textrm{Poisson}(\nu \, \textrm{Leb}(A))$. While the GH family is not convolution-closed in general, it contains two subfamilies that are: the Normal-inverse Gaussian (NIG) distribution, corresponding to $\lambda = -1/2$, and the Variance Gamma (VG) distribution, for $\delta = 0$. Indeed, if $L' \sim \textrm{NIG}(\alpha,\beta,\delta,\mu)$, then $L(A) \sim \textrm{NIG}\left(\alpha,\beta,\delta \, \mathrm{Leb}\left(A\right),\mu\,  \mathrm{Leb}\left(A\right)\right)$ and if $L^{'}\sim \textrm{VG}(\alpha,\beta,\lambda,\mu)$, then $L(A) \sim \textrm{VG}\left(\alpha,\beta, \lambda \,\mathrm{Leb}(A), \mu \, \mathrm{Leb}(A)\right)$. 
These classes preserve closed-form marginals while allowing for both exponential and light tails. A heavy-tailed example is $L^{'} \sim \textrm{Cauchy}(\gamma)$, which yields $L(A) \sim \textrm{Cauchy}(\gamma \cdot \textrm{Leb}(A))$.
 Parameterizations for all probability distributions used in this paper are provided in Section~\ref{subsubsection:parameterizations} and~\ref{subsubsection:alternative_specification}.

\subsection{Kernel convolutions and trawl processes}\label{subsection:trawl_background} 
A general method for constructing stochastic processes $X$ driven by L\'evy bases is based on kernel convolutions. To demonstrate our methodology, we focus on the univariate, time-only case. However, extensions to multivariate, spatio-temporal processes are straightforward \citep{ambit_book,opitz2017spatial}. Under mild regularity conditions on the kernel $K$, the convolution 
\begin{equation*}
    X(\cdot) = \int_{\mathbb{R}^2} K(s-\cdot, y) L(\mathrm{d}s, \mathrm{d}y)
\end{equation*} 
is well-defined \citep[Theorem 2.7]{rajput1989spectral}. Of particular interest are indicator kernel functions, yielding the class of univariate trawl processes, originally introduced by \citet{taqqu} and \cite{barndorff2014integer}. Trawl processes are infinitely divisible, and their marginals often belong to the same family as the L\'evy seed. Formally, consider the collection of sets
\begin{equation*}
    A = \{(s,y) \in \R^2 \colon  s < 0, 0 <  y < a(s) \} , \qquad A_t = A + (t,0),
\end{equation*}
where $a \colon (-\infty,0] \to \mathbb{R}^{+}$ is a smooth, increasing function, and let the kernels be indicators over the sets $A_t$, i.e., $K(s-t,y) = \mathds{1}\left({(s-t,y) \in A}\right) = \mathds{1}\left({(s,y) \in A_t}\right)$. Then the trawl process $X$ at time $t$, denoted by $X_t \equiv X(t)$, is given by the L\'evy basis $L$ evaluated over the trawl set $A_t$, i.e., $X_t = L(A_t)$. A key property of trawl processes is that their statistical properties are fully determined by the autocorrelation structure and marginal distribution. Therefore, trawl processes can be viewed as the natural extension of Gaussian processes to the infinitely divisible setting.

The flexibility of the marginal law of the trawl process is inherited from the L\'evy basis. To maintain a flexible autocorrelation structure, the trawl sets incorporate an additional, abstract dimension $y$. By smoothing the L\'evy basis in higher dimensions, trawl processes can achieve any smooth, decreasing autocorrelation function \citep{barndorff2014integer}. 
Precisely, their correlation structure is given by 
\begin{equation*}
    \rho(h)  \defeq \Corr(X_t,X_{t+h}) = \frac{\mathrm{Leb}\left(A \cap A_h\right)}{\mathrm{Leb}\left(A\right)} = \frac{\int_{-h}^0 a(s) \mathrm{d}s}{\int_{-\infty}^0 a(s) \mathrm{d}s}, \text{ for }h \ge 0.
\end{equation*}
The simplest parametric example is the exponential trawl function, where $a(s) = e^{\lambda s}$ for $s \le 0,  \lambda >0$, which yields $\rho(h) = e^{-\lambda h}$ for $h \ge 0$. To interpolate between short and long memory, consider the randomized mixture of exponentials $a(s) = \int_0^\infty e^{-\lambda s} \pi(\mathrm{d}\lambda)$ for $s \leq 0$, where the decay rate $\lambda$ is sampled according to $\pi$, a probability measure on $\R^{+}$. This recovers the purely exponential case with the Dirac measure $\pi = \delta_{\lambda_0}$ for some $\lambda_0>0$, and it also includes mixtures of exponentials for $\pi = \sum_{i=1}^n \lambda_{i} \delta_{\lambda_i}$, with $\sum_{i=1}^n \lambda_i =1$. In the simulations in Section~\ref{section:simulation_study}, we use the Inverse Gaussian (IG) trawl function, where $\pi = \textrm{IG}(\gamma, \eta)$ for $\gamma, \, \eta >0$ and $a(s) = \left(1-2s/\gamma^2\right)^{-1/2} \exp{ \left(\eta\left(1-\sqrt{1 - 2s/ \gamma^2}\right)\right)}$, $s \le 0$, yielding semi-long memory with $\rho(h) = \exp{\left(\eta \left(1 - \sqrt{1 + 2h/\gamma^2}\right) \right)}$, $h \geq 0$. By semi-long memory we mean $\rho(\cdot)$ decays slower than exponential but faster than polynomial; see Section~\ref{subsubsection:parameterizations} for the IG parameterization.

All in all, trawl processes are integer or real-valued, continuous-time stochastic processes that can describe a wide range of possible serial correlation patterns in data and whose properties are fully determined by their marginal distribution and autocorrelation function. As a result of the latter, trawl processes and their spatio-temporal and multivariate extensions can be viewed as generalizations of Gaussian processes.

\subsection{List of probability distribution parameterizations}\label{subsubsection:parameterizations}
 The parameterizations for the probability distributions used in the paper are given in Table~\ref{table:parameterizations}. Notation-wise, let $\Z^{+} = \{0,1,\ldots\}$ , $\R^{+} = (0,\infty)$ and $K_{\lambda}$ be the modified Bessel function of the second kind of order $\lambda$.\\
\begin{table}[t]
\caption{Parameterisations for the probability distributions used in the paper: Poisson, Gamma, inverse Gaussian (IG), Gaussian ($\mathcal{N}$), Generalized hyperbolic (GH), Normal-inverse Gaussian (NIG), and Variance-gamma (VG).}
\begin{tabular}{ |c||c|c|c| }
 \hline
 \multicolumn{4}{|c|}{Distributions and their parameterizations} \\
 \hline \hline
 Distribution& Range & Parameters & Density  \\[5pt]  \hline
 $\textrm{Poisson}(\lambda)$ & $ \Z^{+} $   & $\lambda \in \R^{+}$ &   $\frac{\lambda^x e^{-\lambda}}{x!}$  \\  \hline
  $\textrm{Gamma}(\alpha,\beta)$& $\R^{+}$    & $\alpha, \beta \in \R^{+}$   & $\frac{\beta^\alpha}{\Gamma(\alpha)} x^{\alpha-1} e^{-\beta x}$   \\ \hline
 $\textrm{IG}(\gamma,\delta)$& $\R^{+}$     & $\gamma, \delta \in \R^{+}$ & $\frac{\sqrt{\gamma/\delta}}{2 K_{1/2}(\gamma \delta)} \frac{1}{\sqrt{x}} e^{-\frac{1}{2}\left(\delta^2 x^{-1} + \gamma^2 x \right)}$ \\ \hline
$\mathcal{N}(\mu,\sigma^2)$ & $\R$ & \begin{tabular}{c} $\mu \in \R$ \\$\sigma^2 \in \R^{+}$\end{tabular}
  &  $\frac{1}{\sqrt{2 \pi \sigma^2}}e^{-\frac{(x-\mu)^2}{2\sigma^2}}$ \\ \hline
   $\textrm{GH}(\lambda, \alpha,\beta,\delta,\mu)$& $\R$  & \begin{tabular}{c} $\alpha, \delta \in \R^{+}; \, \beta, \mu, \lambda  \in \R$ \\ $ \gamma \defeq \sqrt{\alpha^2 - \beta^2} \in \R $\end{tabular}   & $\frac{\left(\gamma/\delta\right)^{\lambda}}{\sqrt{2 \pi} K_{\lambda}(\delta \gamma)}\frac{ K_{\lambda-1/2}\left(\alpha   \sqrt{\delta^2 + (x - \mu)^2}\right)}{\left(\sqrt{\delta^2 + (x - \mu)^2}/\alpha \right)^{1/2 - \lambda}}                   e^{\beta  (x - \mu)}$ \\ \hline 
   $\textrm{NIG}(\alpha,\beta,\delta,\mu)$& $\R$  & \begin{tabular}{c} $\alpha, \delta \in \R^{+}; \, \beta, \mu \in \R$ \\ $ \gamma \defeq \sqrt{\alpha^2 - \beta^2} \in \R $\end{tabular}   & $\frac{\alpha  \delta  K_1\left(\alpha   \sqrt{\delta^2 + (x - \mu)^2}\right)}{\pi  \sqrt{\delta^2 + (x - \mu)^2}}                   e^{\delta \gamma + \beta  (x - \mu)}$ \\ \hline 
  $\textrm{VG}(\alpha,\beta,\lambda,\mu)$& $\R$  & \begin{tabular}{c} $\alpha,\lambda \in \R^{+}; \, \beta, \mu \in \R$ \\$ \gamma \defeq \sqrt{\alpha^2 - \beta^2} \in \R $\end{tabular}  &$\frac{\gamma^{2\lambda}|x-\mu|^{\lambda-1/2}K_{\lambda-1/2}(\alpha |x-\mu|)}{\sqrt{\pi}\Gamma{(\lambda)}(2\alpha)^{\lambda-1/2}} e^{\beta(x-\mu)}$\\ \hline
\end{tabular}
\label{table:parameterizations}
\end{table}
\subsection{Alternative specification for the NIG distribution}\label{subsubsection:alternative_specification}

The four-parameter specification $\textrm{NIG}(\alpha,\beta,\delta,\mu)$ is weakly identifiable in the sense that tuples $(\alpha,\beta,\delta,\mu)$ with very different values result in roughly the same distribution. Further, although $\alpha$ and $\beta$ are known as tail heaviness and asymmetry parameters, respectively, they also contribute to the mean and variance of the distribution. We propose a novel three-parameter parameterization $\textrm{NIG}(\tilde{\mu},\tilde{\sigma},\tilde{\beta})$ which is easily identifiable and in which the roles of the parameters are disentangled. 

\begin{figure}[t]
    \centering
\subfloat[Illustration of the weak-identifiability of parameters $(\beta,\gamma)$ in the standardized NIG distribution from \eqref{eq:standardised_nig}, and hence weak-identifiability of the four-parameter NIG specification.]{\label{fig:nig_four_params}\includegraphics[width=0.49\textwidth,height=0.49\textheight,keepaspectratio]{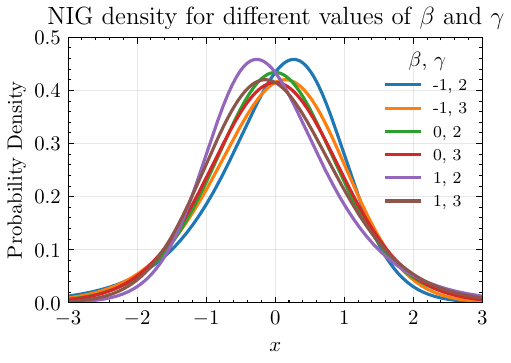}}
\hfill 
\subfloat[$\textrm{NIG}\left(\mu=0,\sigma=1,\beta\right)$ densities at ten equidistant $\beta$ in $(-5,5)$. The densities vary smoothly and $\beta$ is visually identifiable. The legend only shows every third $\beta$ value to avoid clutter.]{\includegraphics[width=0.49\textwidth,height=0.49\textheight,keepaspectratio]{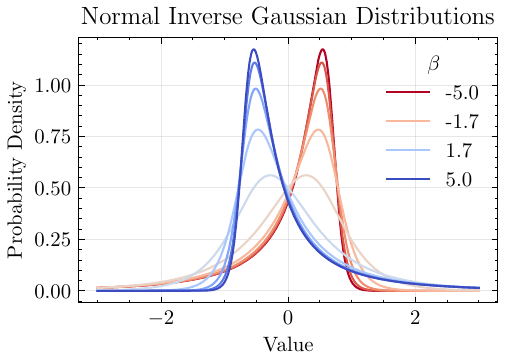}\label{fig:3_param_NIG}}
\caption{Classical and alternative specifications for the NIG distribution.}
\end{figure}

To begin with, note that if $X \sim \textrm{NIG}(\alpha,\beta,\mu,\delta)$, then
\begin{align*}
    \ev\left[X\right] &= \mu +  \delta \beta / \gamma,\\
    \Var\left(X\right)&= \delta \alpha^2 / \gamma^3,
\end{align*}
where $\gamma = \sqrt{\alpha^2-\beta^2}$; we can thus consider the base distribution
\begin{equation} 
    \alpha, \beta \to \textrm{NIG}\left(\alpha = \alpha, \beta = \beta, \delta = \gamma^3/\alpha^2, \mu = -\beta \gamma^2/\alpha^2\right),\label{eq:standardised_nig}
\end{equation}
which is obtained by setting the mean and variance of $X$ to be $0$ and $1$, respectively. Having eliminated the location-scale effects, we can see from Figure~\ref{fig:nig_four_params} that the values of $(\beta,\gamma)$, or equivalently of $(\alpha,\beta)$ are not easily identifiable from the densities. To this end, consider 
$ \gamma(\beta) = 1 + \abs{\beta}/5 $ and consequently $
\alpha(\beta)= \sqrt{\gamma(\beta)^2 + \beta^2} = \sqrt{(1 + \abs{\beta}/5)^2 + \beta^2}$ for $-5 \le \beta \le 5$. Finally, we reintroduce $\tilde{\mu}$ and $\tilde{\sigma}$ by scaling the location and scale parameters of the base distribution, holding $\alpha$ and $\beta$ constant, thus obtaining
\begin{align*}
    (\tilde{\mu},\, \tilde{\sigma}, \, \tilde{\beta}) \to  & \textrm{NIG}\left(\tilde{\mu},\, \tilde{\sigma}, \, \tilde{\beta}\right) \\ \equiv & \textrm{NIG}\left(\alpha = \alpha\left(\tilde{\beta}\right), \beta = \tilde{\beta},\,  \delta = \tilde{\sigma}^2 \,  \gamma^3(\tilde{\beta})/\alpha^2(\tilde{\beta}), \, \mu = \tilde{\mu}-\tilde{\sigma}^2 \,\tilde{\beta}\,  \gamma^2(\tilde{\beta})/\alpha^2(\tilde{\beta})\right),
\end{align*}
which has mean $\tilde{\mu}$, standard deviation $\tilde{\sigma}$ and tilt controlled by $\tilde{\beta}$. Because the two $\textrm{NIG}$ specifications in the above equation have different numbers of parameters, there is no risk of confusion and we drop the tilde and write $\textrm{NIG}\left(\mu,\sigma,\beta\right)$; see Figure~\ref{fig:3_param_NIG} for an illustration of the densities of the new parameterization as a function of $\beta$.

In summary, the two NIG parameterizations are not equivalent. Although the three-parameter one is slightly less flexible, it is more identifiable and parsimonious.

\newpage

\newpage
\section{Further considerations on the TRE}
\label{appendix:subsection_tre}We discussed in Section~\ref{section:meth_contr_to_nre} how the ratio $r{(\x,\bt)} = \dfrac{q(\x,\bt)}{\tilde{q}(\x,\bt)} = \dfrac{p(\x,\bt)}{p(\x)p(\bt)}$ can be approximated by training $m$ classifiers and using the decomposition
\begin{equation*}
r(\x,\bt) = r_m(\x,\bt) \, r_{m-1}(\x,\bt) \ldots  r_1(\x,\bt).
\end{equation*}
The $i^{\text{th}}$ classifier $c_i$ distinguishes between samples from $q_i$ and $q_{i-1}$, where 
\begin{equation}
q_i(\x,\bt) =p(\x,\theta^1,\ldots,\theta^{i}) p(\theta^{i+1},\ldots,\theta^m \mid \theta^1,\ldots,\theta^i) = p(\x,\bt^{1:i}) p(\bt^{i+1:m} \mid \bt^{1:i}),  
\end{equation}
for $ i = 1 , \ldots, m-1$, with $q_m(\x,\bt) = p(\x,\bt)$ and $q_0(\x,\bt) =  p(\x)p(\bt)$. Consequently, the ratios 
\begin{equation}
r_i(\x,\bt) =  \dfrac{q_i(\x,\bt)}{q_{i-1}(\x,\bt)} = 
\dfrac{p(\x,\bt^{1:i}) p(\bt^{i+1:m} \mid \bt^{1:i})}{p(\x,\bt^{1:i-1}) p(\bt^{i:m} \mid \bt^{1:i-1} )} = \dfrac{p(\theta^{i} \mid \x,\bt^{1:i-1})}{p(\theta^{i} \mid \bt^{1:i-1})}\label{eq:TRE_ratio_r_i_supp}
\end{equation}
for $ i = 1 , \ldots, m$. The corresponding BCE loss functions are given by
\begin{equation*}
        \mathcal{L}_i = -\ev_{q_i(\x,\bt)}[ \log{(c_{i}(\x, \bt))}] -\ev_{q_{i-1}(\x,\bt)}[\log{(1- c_{i}(\x,\bt))}], \quad i = 1,\ldots, m,
\end{equation*}
and can be simplified further. By \eqref{eq:TRE_ratio_r_i_supp}, $c_i$ only performs computations using the first $i$ coordinates of $\bt$, i.e., $\bt^{1:i}$. Thus we can interpret $c_i$ as distinguishing between samples from $p(\x,\bt^{1:i})$ and $p(\x,\bt^{1:i-1}) p(\theta^i \mid \bt^{1:i-1})$, and $\mathcal{L}_i$ is equivalent to
\begin{equation*}
    \mathcal{L}_i = -\ev_{p(\x,\bt^{1:i})}[ \log{(c_{i}(\x, \bt^{1:i}))}] -\ev_{p(\x, \bt^{1:i-1})p(\theta^i \mid \bt^{1:i-1})}[\log{(1- c_{i}(\x,\bt^{1:i}))}].
\end{equation*}
The equivalence can also be checked by direct calculation.

We proceed as follows. 
We outline pseudo-code for the TRE training procedure and discuss two relevant neural network architectures in~\ref{subsection:TRE_training_procedure},  prove Theorems~\ref{thm:sample_efficiency} in~\ref{subsection:proof_of_proposition}, and discuss the effect of changing the coordinate ordering in~\ref{subsec:TRE_ordering_SM}. 
\subsection{Training procedure pseudo-code and TRE architecture}\label{subsection:TRE_training_procedure}
\begin{figure}[t] 
    \centering
    \begin{subfigure}[t]{0.49\textwidth}
        \centering
        \includegraphics[width=\textwidth]{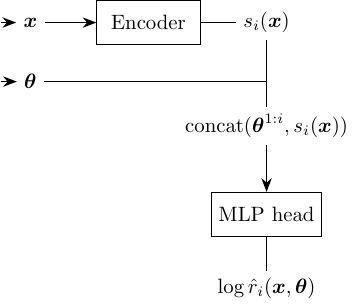}
        \caption{Architecture of $i^{\text{th}}$ classifier within TRE.}
        \label{fig:individual_nre_within_tre}
    \end{subfigure}
    \hfill
    \begin{subfigure}[t]{0.49\textwidth}      
        \centering
        \includegraphics[width=\textwidth]{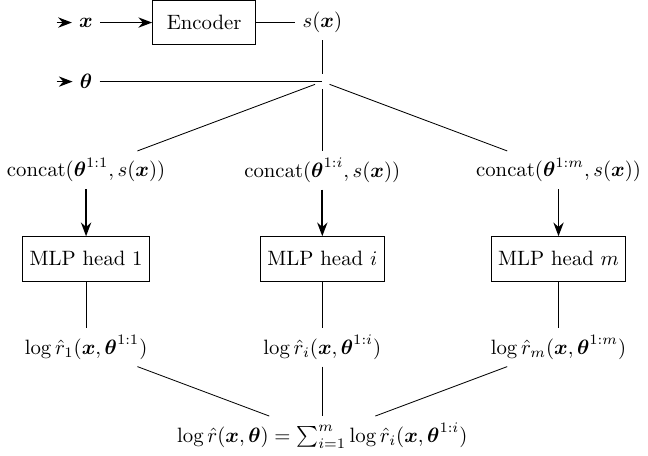}
        \caption{TRE architecture with shared encoder.}
        \label{fig:tre_with_shared_body}
    \end{subfigure}
    \caption{\small{TRE architectures for learning likelihood ratios. Outputs are on the log scale for stability. (a) Independent classifiers: Each encoder (e.g., Long Short-Term Memory, or LSTM) processes the data (e.g., time series) $\x$ into summary statistics $s_i(\x)$, which are concatenated with $\bt^{1:i}$ and passed to a multilayer perceptron (MLP) to approximate 
    $\log \hat{r}_i(\x,\bt)$. 
    (b) Shared encoder variant: All classifiers share a single encoder with separate MLP heads. 
    }} \label{fig:1}
\end{figure}

Finally, we give the pseudo-code for the training procedures from Section~\ref{subsection:NRE_learning_likelihood_through_classification}
\begin{itemize}
    \item NRE in the general setting in Algorithm \ref{algo:general_nre}; here we approximate $r(\z) = \dfrac{q(\z)}{\tilde{q}(\z)}.$ 
    \item NRE specialized to SBI in Algorithm \ref{algo:specialised_nre}; here we approximate \newline $r(\x,\bt) = \dfrac{q(\x,\bt)}{\tilde{q}(\x,\bt)} = \dfrac{p(\x,\bt)}{p(\x)p(\bt)}.$ In this setting, training requires samples from both the joint distribution $p(\x,\bt)$ and the product of marginals $p(\x)p(\bt)$. Samples from $p(\x)p(\bt)$ can be obtained directly from joint samples by shifting: given $(\x_1,\bt_1), (\x_2,\bt_2) \sim p(\x,\bt)$, one forms $(\x_1,\bt_2), (\x_2,\bt_1) \sim p(\x)p(\bt)$, as in Steps~3 and 4 of Algorithm~\ref{algo:specialised_nre}. 
    \item individual classifiers within TRE, specialised to SBI in Algorithm \ref{algo:specialized_tre}; here we approximate each ratio $r_i(\x,\bt) = 
 \dfrac{p(\theta^{i} \mid \x,\bt^{1:i-1})}{p(\theta^{i} \mid \bt^{1:i-1})}$ separately, where $i=1,\ldots,m$. Note that, similarly to the above item, only samples from the joint $p(\x,\bt)$ are required. Further, one such sample can be used for all the $m$ classifiers, simply by dropping $\bt^{i+1:m}$ from $\bt$ and only using $(\x,\bt^{1:i})$ to the component TRE classifiers, for $i=1,\ldots,m$.
\end{itemize}
It is worth discussing further the two main architectural implementations of TRE, illustrated in Figure~\ref{fig:1}. In the first, which we adopt in this paper, the $m$ classifiers are trained separately; see Figure~\ref{fig:individual_nre_within_tre}. Each classifier uses its own encoder, which in our paper is chosen to be a recurrent neural network (RNN), specifically a Long Short-Term Memory (LSTM) neural network, which can process time series of possibly different lengths. Each of the $m$ encoders extracts informative summary statistic $s_i(\x)$ from the time series $\x , i= 1,\ldots, m$. These summary statistics are concatenated with the relevant parameter subset $\bt^{1:i}$, then passed through a multi layer perceptron (MLP) and optimized to minimize a finite-sample approximation of $\mathcal{L}_i$ to estimate 
$\log \hat{r}_i(\x,\bt)$. 

An alternative design employs a shared encoder across all classifiers, as in Figure~\ref{fig:tre_with_shared_body}. In this case, a single encoder produces one shared latent representation $s(\x)$, which is then fed to $m$ separate MLP heads, each corresponding to one intermediate ratio $r_i$. 

Finally, note that the same samples $\{(\x,\bt)\}_{i=1}^N$ can be used for all individual TRE classifiers, regardless of whether the encoder is shared or not. Thus, the individual classifiers from Figure~\ref{fig:individual_nre_within_tre}, or the corresponding one from \ref{fig:tre_with_shared_body} do not require multiple datasets.
\begin{algorithm}[t]
        \caption{Neural ratio estimation, general version.}\label{algo:general_nre}
  \begin{algorithmic}[1]
      \Require Samplers for $q(\z)$ and $\tilde{q}(\z)$, untrained parametric classifier $c_{\pss}(\z)$, learning rate $\lambda$, number of samples $N$, number of training iterations $n$.
    \Ensure Approximations $\hat{c}\left(\z \right)$ of $c^{*}\left(\z \right)$ and $\hat{r}(\z)$ of $r(\z) = \frac{q(\z)}{\tilde{q}(\z)}$.
    \For{$\text{iter}\in \{1,\ldots,n\}$}
    \State Generate samples $\z_j \stackrel{\text{iid}}{\sim} q(\z)$ and $\tilde{\z}_j \stackrel{\text{iid}}{\sim} \tilde{q}(\z)$ for $1 \le j \le N$.
     \State $\mathcal{D} \gets \{\z_1,\ldots,\z_N\}$
     \State $\tilde{\mathcal{D}} \gets \{\tilde{\z}_1,\ldots,\tilde{\z}_N\}$
       \State $\mathcal{L(\pss)} \gets -\displaystyle \frac{1}{2N}\left(\sum_{\z \in \mathcal{D}}\log{\left(c_{\pss}(\z)\right)} + \sum_{\tilde{\z} \in \tilde{\mathcal{D}}}\log(1-c_{\pss}(\tilde{\z}))\right)$ 
    \State $\pss \gets \pss - \lambda \nabla{\mathcal{L}}(\pss)$ \Comment{gradient descent step}
      \EndFor
    \Return $c_{\pss}$
    \end{algorithmic}
\end{algorithm}
\begin{algorithm}[t]
        \caption{Neural ratio estimation, specialised for SBI.}
  \begin{algorithmic}[1]
      \Require Samplers for $\bt \to p(\bt)$ and $\x \to p(\x \mid \bt)$, untrained parametric classifier $c_{\pss}(\x,\bt)$, learning rate $\lambda$, number of samples $N$, number of training iterations $n$.
    \Ensure Approximations $\hat{c}\left( \x,\bt \right)$ of $c^{*}\left(\x,\bt \right)$ and $\hat{r}(\x,\bt)$ of $r(\x,\bt) = \dfrac{p(\x,\bt)}{p(\x)p(\bt)}$.
    \For{$\text{iter} \in \{1,\ldots,n\}$}
    \State Generate samples $\bt_j \stackrel{\text{iid}}{\sim} p(\bt)$ and $\x_j\vert  \stackrel{\text{iid}}{\sim} p(\x \vert \bt_j)$ for $1 \le j \le N$.
    \State $\mathcal{D} \gets \{(\x_1,\bt_1),\ldots,(\x_N,\bt_N)\}$ \Comment{samples from the joint density $p(\x,\bt)$}
    \State $\tilde{\mathcal{D}} \gets \{(\x_1,\bt_2),\ldots,(\x_{N-1},\bt_N), (\x_N,\bt_1)\}$ \Comment{\parbox[t]{.445\linewidth}{shift joint samples to get samples from the product of marginals $p(\bt)p(\x)$}} 
    \State $\mathcal{L}(\pss) \gets \displaystyle -\frac{1}{2N}\left(\sum_{(\x,\bt) \in \mathcal{D}}\log(c_{\pss}(\x,\bt)) + \sum_{(\x,\bt) \in \tilde{\mathcal{D}}}\log{\left(1-c_{\pss}(\x,\bt)\right)}\right)$ \label{algo:specialized_nre}
    \State $\pss \gets \pss - \lambda \nabla{\mathcal{L}}(\pss)$ \Comment{gradient descent step}
      \EndFor
    \Return $c_{\pss}$
    \end{algorithmic}
    \label{algo:specialised_nre}
\end{algorithm}
\begin{algorithm}[t]
        \caption{Telescoping ratio estimation ($i^\text{th}$ classifier), specialised for SBI.}
  \begin{algorithmic}[1]
      \Require Samplers for $\bt \to p(\bt)$, $\theta^i \to p(\theta^i \mid \bt^{1:i-1})$ and $\x \to p(\x \mid \bt)$, untrained parametric classifier $c_{i,\pss}(\x,\bt^{1:i})$, learning rate $\lambda$, number of samples $N$, number of training iterations $n$.
    \Ensure Approximations $\hat{c}_i\left( \x,\bt \right)$ of $c^{*}_i\left(\x,\bt \right)$ and $\hat{r}_i(\x,\bt)$ of \newline $   r_i(\x,\bt) =  \dfrac{q_i(\x,\bt)}{q_{i-1}(\x,\bt)}  = \dfrac{p\left(\theta^{i} \mid \x,\bt^{1:i-1}\right)}{p\left(\theta^{i} \mid \bt^{1:i-1}\right)}$
    .
    \For{$\text{iter} \in \{1,\ldots,n\}$}
    \State Generate samples $\bt_j \stackrel{\text{iid}}{\sim} p(\bt)$ and $\x_j \stackrel{\text{iid}}{\sim} p(\x \vert \bt_j)$ for $1 \le j \le N$.
    \State Generate $\tilde{\theta}_j^i\sim p(\theta^i \mid \bt_j^{1:i-1})$ for $1 \le j \le N$.
    \State $\mathcal{D} \gets \{(\x_1,\bt_1^{1:i}),\ldots,(\x_{N-1},\bt_{N-1}^{1:i}), (\x_N,\bt_N^{1:i})\}$  \Comment{samples from $q_i(\x,\bt)$}
    \State $ \tilde{\mathcal{D}} \gets \{(\x_1,(\bt_1^{1:i-1},\tilde{\theta}_2^{i})),\ldots,(\x_N,(\bt_N^{1:i-1},\tilde{\theta}_1^{i}))\}$ \Comment{samples from $q_{i-1}(\x, \bt)$}
    \State $\mathcal{L}_i(\pss) \gets -\displaystyle \frac{1}{2N}\left(\sum_{(\x,\bt) \in \mathcal{D}}\log(c_{i,\pss}(\x,\bt)) + \sum_{(\x,\bt) \in \tilde{\mathcal{D}}}\log{\left(1-c_{i,\pss}(\x,\bt)\right)}\right)$ 
    \State $\pss \gets \pss - \lambda \nabla{\mathcal{L}_i}(\pss)$ \Comment{gradient descent step}
      \EndFor
    \Return $c_{i,\pss}$
    \end{algorithmic}
    \label{algo:specialized_tre}
\end{algorithm}

\subsection{Proof of Theorem~\ref{prop:kl_decomposition} }\label{subsection:proof_of_proposition}
To begin with, we have that 
\begin{align}
    & \sum_{i=1}^m \kl{q_i}{q_{i-1}} = \sum_{i=1}^m \int q_i(\x,\bt) \log{\frac{q_i(\x,\bt)}{q_{i-1}(\x,\bt)}} \, \mathrm{d}\x \mathrm{d}\bt \notag\\ 
    =& \sum_{i=1}^m \int p(\x,\bt^{1:i}) p(\bt^{i+1:m} \mid \bt^{1:i} )\log{\frac{p(\x,\bt^{1:i})p(\bt^{i+1:m}\mid \bt^{1:i})}{p(\x,\bt^{1:i-1})p(\bt^{i:m} \mid \bt^{1:i-1})}} 
    \, \mathrm{d}\x \mathrm{d}\bt  \notag \\
    =& \sum_{i=1}^m \int p(\x,\bt^{1:i}) p(\bt^{i+1:m} \mid \bt^{1:i}) \log{\frac{p(\theta^i \vert \x,\bt^{1:i-1})}{p(\theta^i \vert \bt^{1:i-1} )}} \, \mathrm{d}\x \mathrm{d}\bt \notag \\
    =& \sum_{i=1}^m \int p(\x,\bt^{1:i})  \log{\frac{p(\theta^i \vert \x,\bt^{1:i-1})}{p(\theta^i \vert \bt^{1:i-1} )}} \, \mathrm{d}\x \mathrm{d}\bt^{1:i}   \label{eq:marginal_1} \\
     =&  \sum_{i=1}^m \int p(\x,\bt^{1:i})  \log{p(\theta^i \vert \x,\bt^{1:i-1})} - p(\x,\bt^{1:i})  \log{p(\theta^i \vert \bt^{1:i-1} )} \, \mathrm{d}\x \mathrm{d}\bt^{1:i}  \notag\\
         =&  \sum_{i=1}^m \int p(\x,\bt^{1:i})  \log{p(\theta^i \vert \x,\bt^{1:i-1})}  \mathrm{d}\x \mathrm{d}\bt^{1:i}  - \sum_{i=1}^m \int p(\bt^{1:i})  \log{p(\theta^i \vert \bt^{1:i-1} )} \, \mathrm{d}\x \mathrm{d}\bt^{1:i}  \label{eq:sum_KL}
    \end{align}
where \eqref{eq:marginal_1} and \eqref{eq:sum_KL} follow by marginalization. We compute each of the two terms separately. The first term 
simplifies to
\begin{align*}
    \sum_{i=1}^m &\int p(\x,\bt^{1:i}) \log{p(\theta^i \vert \x,\bt^{1:i-1})}\, \mathrm{d}\x \mathrm{d}\bt^{1:i} 
    = \sum_{i=1}^m \int p(\x,\bt ) \log{p(\theta^i \vert \x,\bt^{1:i-1})}  \, \mathrm{d}\x \mathrm{d}\bt \\
    =  &\int p(\x,\bt )  ( \sum_{i=1}^m \log{p(\theta^i \vert \x,\bt^{1:i-1})} )  \, \mathrm{d}\x \mathrm{d}\bt 
    =  \int p(\x,\bt ) \log{p(\bt \vert \x)} \, \mathrm{d}\x \mathrm{d}\bt \\
     = & \int p(\x,\bt ) \log{\frac{p(\bt,\x)}{p(\x)p(\bt)}} \, \mathrm{d}\x \mathrm{d}\bt + \int p(\x,\bt ) \log{p(\bt)} \, \mathrm{d} \x \mathrm{d}\bt = \kl{q_m}{q_0} + \int p(\bt ) \log{p(\bt)} \, \mathrm{d}\bt \\
     = &  \kl{q_m}{q_0} - \mathcal{H}(p\left(\bt\right)), 
\end{align*}
where $\mathcal{H}(\cdot)$ denotes the entropy of a given density. The second term simplifies to 
\begin{align*}
     \sum_{i=1}^m \int p(\bt^{1:i})  \log{p(\theta^i \vert \bt^{1:i-1} )} \, \mathrm{d}\bt^{1:i} =&  \sum_{i=1}^m   \int p(\bt)  \log{p(\theta^i \vert \bt^{1:i-1} )} \,  \mathrm{d}\bt = \int p(\bt) \sum_{i=1}^m  \log{p(\theta^i \vert \bt^{1:i-1} )} \, \mathrm{d}\bt \\
   = &  \int p(\bt)  \log{p(\bt)} \mathrm{d}\bt = - \mathcal{H}(p(\bt)).
\end{align*}
Plugging these expressions back into \eqref{eq:sum_KL} gives
\begin{equation*}
\boxed{\kl{q_m}{q_0} = \sum_{i=1}^m \kl{q_i}{q_{i-1}}.}
\end{equation*}
\subsection{Changing the order of components in the TRE}\label{subsec:TRE_ordering_SM}
Let $1 \le r  < s < n < m$ and define the index sets 
\begin{align*}
    r+1:m\backslash\{n\} = \{r+1,\ldots,n-1,n+1,m\},\\
    s+1:m\backslash\{n\} = \{s+1,\ldots,n-1,n+1,m\}.
\end{align*}
Consider the following two $\mathcal{S}$ values
\begin{align}
    \mathcal{S}_{n \mid 1:r} = \kl{p(\x,\bt^{1:r},\theta^n)p(\bt^{r+1:m\backslash\{n\}} \mid \bt^{1:r},\theta^n)}{p(\x,\bt^{1:r})p(\bt^{r+1:m} \mid \bt^{1:r})}&, \label{eq:S_n_1_r} \\
      \mathcal{S}_{n \mid 1:s} = \kl{p(\x,\bt^{1:s},\theta^n)p(\bt^{s+1:m\backslash\{n\}} \mid \bt^{1:s},\theta^n)}{p(\x,\bt^{1:s})p(\bt^{s+1:m} \mid \bt^{1:s})}& \notag,
\end{align}
which correspond to learning $\theta^n$ conditional on $\bt^{1:r} = (\theta^1,\ldots,\theta^r)$ and $\bt^{1:s} = (\theta^1,\ldots,\theta^r,\ldots,\theta^s)$, respectively. Intuitively, these correspond to the coordinate orderings $\theta^1,\ldots,\theta^r, \theta^n $ and $\theta^1,\ldots,\theta^r,\ldots,\theta^s,\theta^n$. Note that the ratio of the two densities in $\kl{\cdot}{\cdot}$ from \eqref{eq:S_n_1_r} is 
\begin{equation*}
    \frac{p(\x,\bt^{1:r},\theta^n)p(\bt^{r+1:m\backslash\{n\}} \mid \bt^{1:r},\theta^n)}{p(\x,\bt^{1:r})p(\bt^{r+1:m} \mid \bt^{1:r})} = \frac{p(\theta^n \mid \x, \bt^{1:r})}{p(\theta^n \mid \bt^{1:r})}.
\end{equation*}
A simple calculation shows that
\begin{align*}
  \mathcal{S}_{n \mid 1:r} =& \int p(\x,\bt^{1:r},\theta^n)p(\bt^{r+1:m\backslash\{n\}} \mid \bt^{1:r},\theta^n) \log \frac{p(\theta^n \mid \x, \bt^{1:r})}{p(\theta^n \mid \bt^{1:r})} \mathrm{d}\x \, \mathrm{d}\bt^{1:r} \, \mathrm{d}\bt^{r+1:m}   \\ 
  =& \int p(\x,\bt^{1:r},\theta^n)\log \frac{p(\theta^n \mid \x, \bt^{1:r})}{p(\theta^n \mid \bt^{1:r})} \mathrm{d}\x \, \mathrm{d}\bt^{1:r} \, \mathrm{d}\theta^n.
\end{align*}
By an analogous argument
\begin{align*}
  \mathcal{S}_{n \mid 1:s} =&  \int p(\x,\bt^{1:s},\theta^n)\log \frac{p(\theta^n \mid \x, \bt^{1:s})}{p(\theta^n \mid \bt^{1:s})} \mathrm{d}\x \, \mathrm{d}\bt^{1:s}\, \mathrm{d}\theta^n.
\end{align*}
By taking the difference, we recognize the identity 
\begin{equation}
    \mathcal{S}_{n \mid 1:s} - \mathcal{S}_{n \mid 1:r} = I(\theta^n; \bt^{r+1:s} \mid \x, \bt^{1:r}) - I(\theta^n; \bt^{r+1:s} \mid \bt^{1:r}),
    \label{eq:cond_mutual_info_diff}
\end{equation}
where the mutual information of random variables $\boldsymbol{X}$ and $\boldsymbol{Y}$ conditional on $\boldsymbol{Z}$ is given by
\begin{equation*}
    I(\boldsymbol{X};\boldsymbol{Y} \mid \boldsymbol{Z}) = \int p(\x,\y,\z) \log{\frac{p(\x \mid \y,\z)}{p(\x\mid \z)}} \mathrm{d} \x \, \mathrm{d} \y\, \mathrm{d} \z.
\end{equation*}
It is well known that the conditional mutual information is always non-negative. If the sampling distribution factorizes as $    p(\bt) = p(\theta^1) \cdots p(\theta^m)$, then the second term from~\eqref{eq:cond_mutual_info_diff} is $0$, and it follows that
\begin{equation*}
     \mathcal{S}_{n \mid 1:s} - \mathcal{S}_{n \mid 1:r} = I(\theta^n; \bt^{r+1:s} \mid \x, \bt^{1:r}) .
\end{equation*}
In this case, the difference in $\mathcal{S}$ values is precisely given by a conditional mutual information which accounts for the extra information above from the added conditioning coordinates $\theta^{r+1},\ldots,\theta^{s}$.
\newpage
\section{Sequential posterior sampling with Chebyshev polynomials in the TRE framework}
\label{suppl:subsection_chebyshev}
In Section~\ref{sec-meth}, we showed that directly approximating the likelihood ratio $r(\x,\bt) = \dfrac{p(\x,\bt)}{p(\x)p(\bt)}$ becomes increasingly difficult as the number of parameter $m$ grows, limiting the applicability of NRE to complex statistical models. We addressed this issue by introducing the interpolating densities $q_0,\ldots,q_m$ with $q_i(\x,\bt) = p\left(\x,\bt^{1:i}\right) p\left(\bt^{i+1:m}\right)$, $q_m(\x,\bt) = p(\x,\bt)$ and $q_0(\x,\bt) = p(\x)p(\bt)$
and training $m$ classifiers, each of which learns a one-dimensional conditional density of the form $p(\theta^i \vert \x, \bt^{1:i-1})$. As explained before, this approach gives classifiers with better finite sample properties. Further, it allows for efficient posterior sampling, limiting the need for MCMC methods or bypassing MCMC entirely. Once $\hat{p}(\theta^i \vert \x, \bt^{1:i-1})$ is available from an individual classifier within the TRE framework, we can construct a computationally efficient density approximation $\hat{p}_{\text{Cheb}}(\theta^i \vert \x, \bt^{1:i-1})$ using Chebyshev polynomials, which enables fast sampling, computation of highest density regions, and much more. While this introduces a second layer of approximation (namely, approximating the TRE output, which itself approximates the true density), it crucially decouples expensive neural network evaluations from the statistical diagnostics required for validating the SBI model through checks. By contrast, each new MCMC iteration requires new classifier evaluations. To fully understand the benefits of our MCMC-free approach, we first introduce the reader to the ideas of approximating a probability density function by Chebyshev polynomials, and then discuss in Section~\ref{sm:per_parmaeter_checks} and \ref{SM:seq_p_sam} the exact applicability to the individual coverage checks and sequential posterior sampling in the TRE framework, respectively. We also discuss the case of learning blocks of two parameters. Finally, we stress that approximating one-dimensional densities is far less computationally expensive than approximating two-dimensional densities. Extending beyond the bivariate case is an active research area \citep{chebfun_3d}.
\subsection{Chebyshev polynomials background}\label{SM:cheb_background}
The Chebyshev polynomials of the first kind $T_n :[-1,1] \to \R$ are formally defined as 
\begin{equation*}
    T_n(x) \;=\; \cos\left(n \arccos x\right), \ n = 0,1,2,\dots
\end{equation*}
Equivalently, they satisfy the three‐term recurrence:
\begin{align}
    T_0(x) &= 1, \notag \\ 
    T_1(x) &= x, \notag \\
    T_{n+1}(x) &= 2x\,T_n(x) \;-\; T_{n-1}(x), \label{eq:3_term_recurrence}
\end{align}
and are orthogonal with respect to the weight function $w: [-1,1] \to \R, w(x) = \frac{1}{\sqrt{1-x^2}},$ i.e.
\begin{equation*}
        \int_{-1}^{1} T_m(x)\,T_n(x)\,w(x)\,dx
    = 
    \begin{cases}
      0, & m \neq n,\\
      \pi, & m = n = 0,\\
      \tfrac{\pi}{2}, & m = n \ge 1.
    \end{cases}
\end{equation*}
Let $f:[-1,1] \to \R$ be a continuous function. By basic approximation results in Hilbert spaces, there exist coefficients $a_0, a_1, \ldots$ such that
\begin{equation*}
f(x) = \sum_{n=0}^\infty a_n T_n(x) \text { for any } x \in [-1,1].
\end{equation*}
It is a well known fact that approximating a function by high degree polynomials can be unstable, especially close to the endpoints, e.g., see Runge's phenomenon. Instead of interpolating at equidistant points in $[-1,1]$, Chebyshev polynomials are interpolated at the Chebyshev knots $\cos{\left(\dfrac{k \pi}{n}\right)}$ for $k =0, \ldots, n$, which cluster near the endpoints and alleviate the oscillations. If $f$ is absolutely continuous, then the truncation $S_N(f) \defeq \sum_{n=0}^N a_n T_n$ converges uniformly to $f$, and further if $f$ is $n+1$ times continuously differentiable, then
\begin{equation*}
    \|f- S_n(f)\|_{\infty} \le \frac{2^n \cdot \|f^{n+1}\|_{\infty}}{n!},
\end{equation*}
where $\|\cdot \|_{\infty }$ is the supremum norm on $[0,1]$. By contract, for interpolation at equispaced nodes, the upper bound behaves like $2^n \cdot \|f^{n+1}\|_{\infty}$, triggering oscillations. Further properties, including geometric convergence for entire or smooth $f$ can be found in \cite{Trefethen}. 

While a given function $f$ can be interpolated at the Chebyshev knots using families of polynomials other than the Chebyshev one, the latter provide practical advantages, out of which we mention
\begin{itemize}
    \item Fast and stable fitting of the coefficients $a_0,\ldots,a_N$ of the truncation $S_n(f) = \sum_{n=0}^N a_n T_n \approx f$ with $\mathcal{O}(N \log N)$ operations using the Fast Fourier Transform (FFT) to implement the Discrete Cosine Transform (DCT). 
    \item Fast coefficient recurrence for the anti-derivative of $S_n(f)$ with $\mathcal{O}(N)$ operations. Based on the equation
    \begin{equation*}
\int T_k(x)\,\mathrm{d}x = \frac{1}{2}\left(
\frac{T_{k+1}(x)}{k+1}-\frac{T_{k-1}(x)}{k-1}\right)\qquad \text{ for any } k \ge 1,
    \end{equation*}
the following two-term recurrence can be derived
\begin{align*}
    \int S_n(f)(x) \, \mathrm{d} x &= \int \sum_{n=0}^N a_n T_n(x) \, \mathrm{d}x = \sum_{n=1}^{N+1} c_n T_n,  \text{ where } \\
    c_{N+1} &= \frac{a_N}{2N+1},\\
    c_k     &= \frac{a_{k-1} - c_{k+2}}{2(k+1)} \text{ for  } k= N, ,\ldots, 1,
\end{align*}
with the convention $a_{-1} = 0$.
\item Fast evaluation of $S_N(f)$ with $5N$ operations, so $\mathcal{O}(N)$, using the three-term recurrence \ref{eq:3_term_recurrence}. An even better method is Clenshaw's algorithm, which only requires $4N$ operations and is numerically stable for large $N$ \citep{clenshaw1955note}.
\end{itemize}
\subsection{Sampling from univariate and bivariate distributions with Chebyshev polynomials}\label{SM:sampling_from_1d_and_2d}
Assume $f$ is a black-box, univariate probability density function, not necessarily normalized. Let $F$ be the corresponding cumulative distribution function and further $U$ be uniformly distributed on $[0,1]$. The key takeaway from the previous section is that given the values of $f$ at the Chebyshev knots, we can construct an approximation $\tilde{f} \approx f$, then approximate the CDF $\tilde{F} \approx F$ and then sample $X = \tilde{F}^{-1}(U)$ by inversion.  \cite{olver2013fast} explains that it takes only 53 iterations for the bisection method to reach $\textrm{float64}$ precision. Whereas other root finding algorithms, e.g., Newton's method may have better properties, they might take longer to converge for specific values of $U$, preventing efficient parallelization. Note that evaluating the approximate CDF $\tilde{F}$ is extremely fast and also stable when implemented via Clenshaw's algorithm.

\begin{figure}[t]
    \centering
    \includegraphics[]{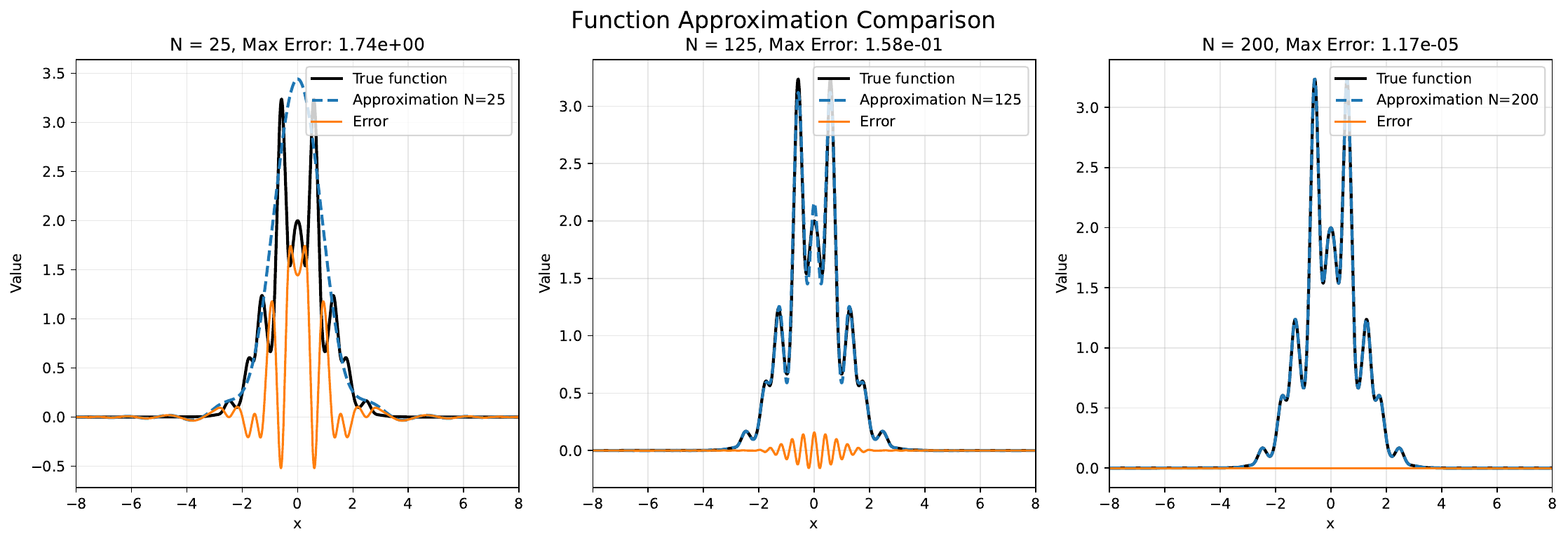}
    \caption{\small{Comparison of the approximations $\tilde{f}_N$ of the unnormalized density $f$ by Chebyshev polynomials of order $N = 25, 125, 200$ (left to right), together with the corresponding approximation error. Note that $N=200$ produces visually indistinguishable results on this oscillating, multimodal function $f$.}}
    \label{fig:sm_cheb_unnorm}
\end{figure}

\begin{figure}[H]
    \centering
    \includegraphics[width=0.49\textwidth]{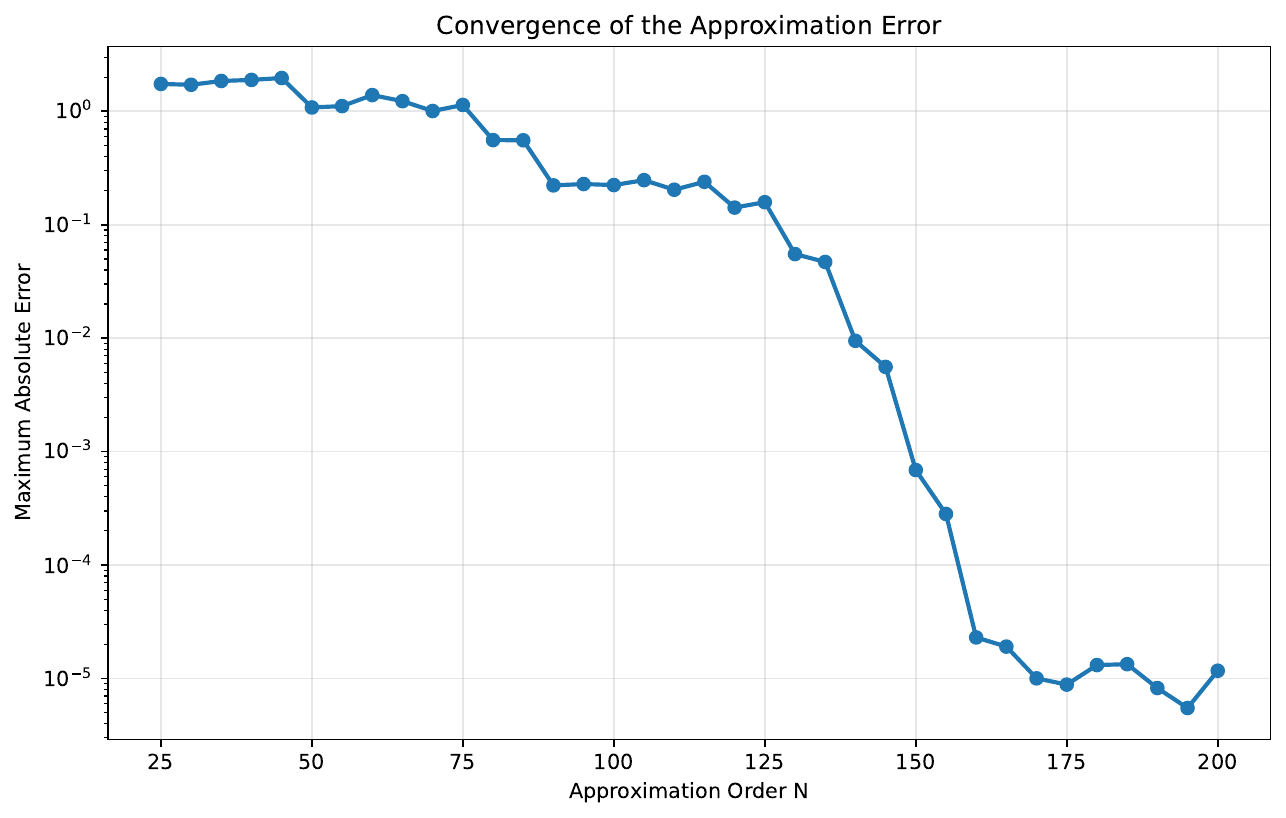}\hfill
    \includegraphics[width=0.49\textwidth]{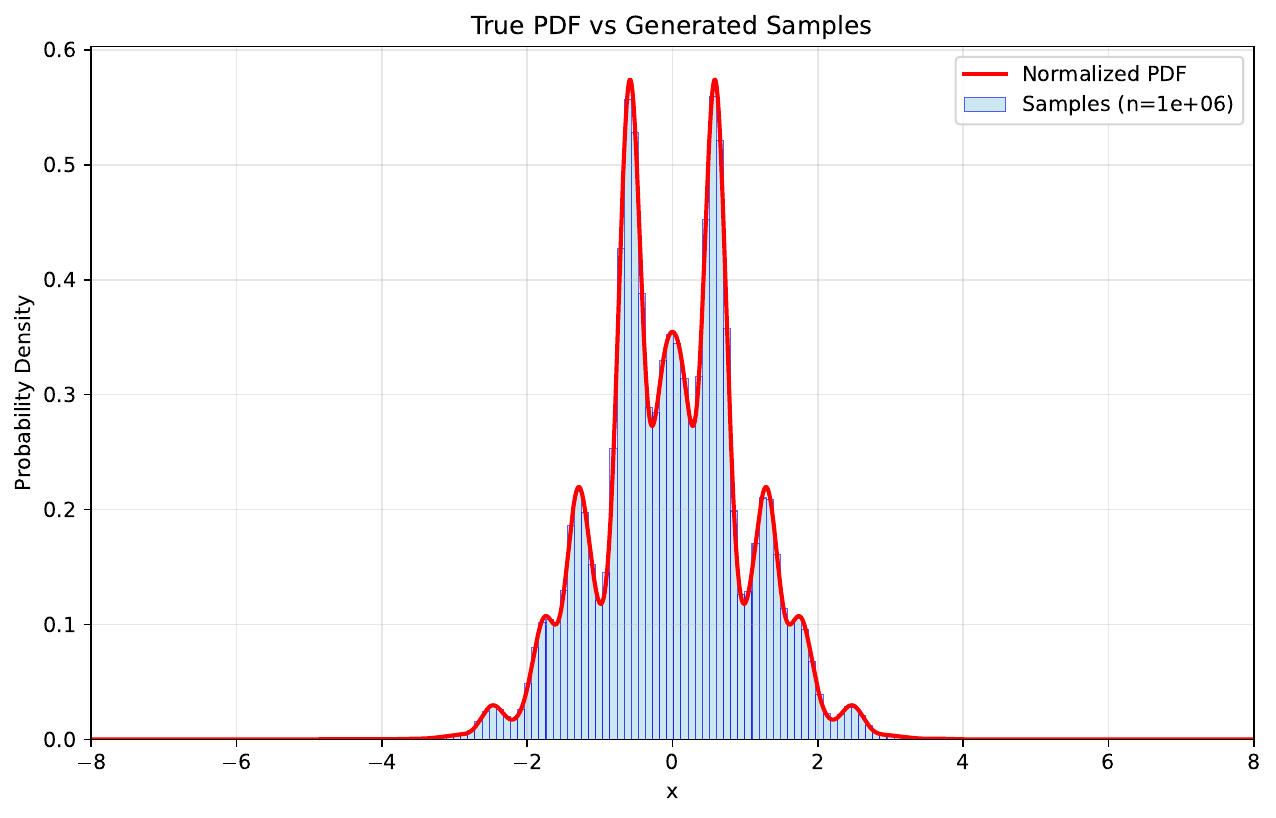}
    \caption{\small{Left: Approximation error $\|f - \tilde{f}_N \|$ as a function of $N$. Right: Comparison between the normalized density $f / \int_{-8}^8 f(u) \mathrm{d}u$  and a histogram based on $10^6$ samples, generated fro the approximation. The results demonstrate that the approximation is highly accurate; in fact, the error for the normalized density is smaller than the error shown in the left panel by a factor of approximately $5.64$, which corresponds to the normalizing constant.}}
    \label{fig:sm_cheb_sidebyside}
\end{figure}

The above technique decouples the expensive evaluations of $f$ from the sampling procedure, by building an approximation $\tilde{f} \approx f$ which can then be used to generate an arbitrary number of samples almost instantaneously, in parallel. We illustrate the approach on
\begin{equation*}
    f(x) = \exp{\left(-\tfrac{x^{2}}{2}\right)} \,\bigl(1 + \sin^{2}(3x)\bigr)\,\bigl(1 + \cos^{2}(5x)\bigr), \quad -8 \le x \le 8.
\end{equation*}
This example is particularly challenging because $f$ is multimodal, oscillates within the interior of the domain, and decays to nearly zero near the boundaries. Figure $\ref{fig:sm_cheb_unnorm}$ shows that the approximation by Chebyshev polynomials $\tilde{f}_N$ converges to $f$ uniformly as $N \to \infty$. Further, Figure~\ref{fig:sm_cheb_sidebyside} shows the uniform error  $\|f - \tilde{f}_N\|$ as a function of $N$, alongside a comparison between the normalized $f$ and a histogram based of $10^6$ samples 
drawn from the Chebyshev approximation with $N=200$. As explained above, once $\tilde{f}_N$ is available, sampling can be done exactly up to machine precision via bisection, as the CDF of the approximation is available in closed form.

\subsubsection{Bivariate distributions and higher dimensions}\label{SM:2d_cheb_case}

The bivariate case can be tackled similarly, by approximating the function $f$ by products of polynomials of the form
\begin{equation*}
    f(x,y) \approx \sum_{i,j} c_j T_i(x) T_j(y),
\end{equation*}
for some chosen pairs $(i,j)$. The same approach goes through, including efficient sampling based on Clenshaw's algorithm; fitting the coefficients requires evaluations of the computationally expensive function $f$ of a two-dimensional grid though, see \cite{olver2013fast}. As expected, the two-dimensional case is more computationally expensive than the one-dimensional case.
\subsection{Per-parameter posterior sampling}\label{sm:per_parmaeter_checks}
Consider the problem of estimating the expected coverage at level $1- \alpha$ for parameter $\theta^i$
\begin{equation*}
    \hat{\mathcal{C}}^{i}_{1-\alpha} = \ev_{p(\x,\bt)}\left[\mathds{1}\left(\theta^i  \in \Theta_{\hat{p}(\vartheta^i  \, \vert \x, \bt^{1:i-1} )}(1-\alpha) \right) \right],\label{eq:from_supl_coverage_def_from_suppl}
\end{equation*}
where $\Theta_{\hat{p}(\vartheta^i \, \vert \x, \bt^{1:i-1} )}(1-\alpha)$ denotes the $1-\alpha$ highest posterior density region (HPD) of $\vartheta^i\mapsto \hat{p}(\vartheta^i \, \vert \x, \bt^{1:i-1} )$, and where we use $\vartheta^i$ instead of $\theta^i$ as argument to avoid confusion. The HPD and $\hat{\mathcal{C}}^{i}_{1-\alpha}$ are approximated as follows: generate $N$ pairs $(\x_j,\bt_j)$, $j=1,\ldots,N$; for each $\x_j$, generate $M$ posterior samples $\vartheta^{i}_{j,1},\ldots,\vartheta^{i}_{j,M} \sim p \left(\vartheta^i \vert \x_j, \bt_j^{1:i-1}\right)$, sort  $p\left(\vartheta^{i}_{j,1} \vert \x_j, \bt^{1:i-1} \right),\ldots,p\left(\vartheta^{i}_{j,M} \vert \x_j, \bt_j^{1:i-1}\right)$ in descending order and take the top $(1-\alpha) M$ of them. The posterior density of the last included sample is then the threshold used to determine acceptance to the approximate HPD $\Theta^M_{\hat{p}(\cdot \, \vert \x_j, \bt_j^{1:i-1} )}(1-\alpha) \approx \Theta_{\hat{p}\left(\cdot \, \vert \x_j, \bt_j^{1:i-1}\right)}(1-\alpha)$. Finally, compare this threshold with $p(\theta^i_j \vert \x_j)$ and calculate the proportion of true parameters that fall within their corresponding HPD regions
\begin{equation}
   \hat{C}^i_{1-\alpha} \approx \sum_{j=1}^N\mathds{1}\left( \bt_j \in \Theta^M_{\hat{p}(\cdot \, \vert \x_j, \bt_j^{1:i-1} )}(1-\alpha)\right) = \sum_{j=1}^N\mathds{1}\left( p(\bt_j,\x_j) \ge \textrm{threshold}_j \right) .\label{eq:coverage_estimation_empirically_S}
\end{equation}
The same procedure is applicable without change when learning blocks of coordinates at once, as we did in Section~\ref{section:simulation_study}.
\subsection{Sequential posterior sampling}\label{SM:seq_p_sam}
The per-parameter posterior sampling described above has a great advantage. The Chebyshev approximation only needs to be built once for each realization$(\bt_j, \x_j)$. By contrast, when performing the posterior sampling check from Section~\ref{subsection:coverage_diagnostics}, we have to 
\begin{itemize}
    \item Construct a Chebyshev approximation for $p(\theta^1 \mid \x_j)$ and draw $M$ samples $\vartheta^1_{j,1},\ldots, \vartheta^1_{j,M}$.
    \item For the second parameter, construct Chebyshev approximations for each of the $M$ conditional densities $p(\theta^2 \mid  \vartheta^1_{j,1}, \x_j)$, $\ldots$, $p(\theta^2 \mid  \vartheta^1_{j,M}, \x_j)$ and generate samples $\vartheta^2_{j,1},\ldots, \vartheta^2_{j,M}$, one from each of the enumerated densities.
    \item Continue sequentially up to the $m^{\text{th}}$ parameter, constructing one Chebyshev approximation for each of the densities $M$ conditional densities $p(\theta^m \mid  \vartheta^{1:M-1}_{j,1}, \x_j)$, $\ldots$, $p(\theta^m \mid  \vartheta^{1:m-1}_{j,M}, \x_j)$ and generate one sample from each $\vartheta^m_{j,1},\ldots, \vartheta^m_{j,M}$.
\end{itemize} 
While the per-parameter checks require a fixed number of neural network evaluations to compute $p(\bt_j, \x_j) \ge \textrm{threshold}_j$ from \eqref{eq:coverage_estimation_empirically_S}, sequential sampling multiplies the number of evaluations by the number of generated samples $M$. Despite this increased cost, the method remains far more computationally efficient than MCMC samplers in our experiments.

Importantly, the performance of this method does not depend on whether the target distribution is multimodal or not, a fact which is known to cause significant problems to MCMC schemes. Moreover, it can be used in conjunction with MCMC. If Chebyshev approximations are not deemed accurate enough, the generated samples can be used as starting point for MCMC, as in particle filters. Since the generated samples are independent, they provide good estimates for the covariance matrix proposal of the MCMC scheme and eliminate, or at least reduce, the need for a burn-in period. All in all, we found that the Chebyshev polynomial approximations perform accurately, reliably, and efficiently, and can also be accelerated on a GPU. 

Overall, Chebyshev polynomial approximations provide an approach that is accurate, reliable, and efficient for posterior sampling, with additional acceleration possible on a GPU. Another advantage is in finding the MAP (MLE): generating a few tens of posterior samples (almost instantaneously) and selecting the point that has the highest posterior (likelihood) gives a strong starting point for numerical maximization algorithms.
\newpage
\section{Calibration and coverage metrics}\label{SM:calibration_and_coverage}
\subsection{Expected calibration error}\label{SM:ECE}
Let $c : \mathcal{Z} \to [0,1]$ be a binary classifier. A popular scalar measure of classifier miscalibration is the expected calibration error (ECE) 
\begin{equation*}
    \ev\left[\left|\mathds{P}\left(Y =1 \mid c(\mathbf{Z}) \right)-c(\mathbf{Z}) \right|\right].
\end{equation*}
The ECE can be approximated from a dataset $\mathcal{D}$ of sample and label pairs $(\z,Y)$, by discretizing the interval $[0,1]$ into $N$ bins $\{B_i\}_{i=1}^N$, and computing, for each bin $B_i$, the empirical accuracy $A_i$ and confidence $C_i$:
\begin{equation*}
    A_i = \frac{1}{|B_i|}\sum_{\z \in \mathcal{D}: \,  c(\z) \in B_i} \mathds{1}{(Y = \hat{y}(\z))}, \qquad
    C_i = \frac{1}{|B_i|}\sum_{\z \in \mathcal{D}: \,  c(\z) \in B_i} c(\z),
\end{equation*}
where $\hat{y}(\z) = \mathds{1}(c(\z) \ge 0.5)$ and $|B_i|$ denotes the number of samples in bin $B_i$. The empirical ECE is then given by 
\begin{equation*}
    \widehat{\textrm{ECE}} = \sum_{i=1}^{N}\frac{|B_i|}{N}\left|A_i - C_i \right|.
\end{equation*}
Unfortunately, the result can be sensitive to the binning strategy:
\begin{itemize}
\item uniform, for which $B_1 = [0,1/N], \ldots, B_N = [1- 1/N,1]$;
   \item adaptive, equal-frequency bins, based on the distribution of the classifier outputs $c(\z)$.
\end{itemize} 
The latter strategy produces better estimates with lower biases, as it concentrates on subintervals of $[0,1]$ where the classifier outputs concentrate. By comparison, uniform binning can consider bins with almost no samples, for which the accuracy and confidence are difficult to estimate. Regardless of the binning strategy, if the classifier tends to output values that are very close to $0$ and $1$, the differences between the bin edges can be close to $0$, posing numerical challenges.
\subsection{Rank checks} \label{SM:ranks}
Next, we consider rank checks, which were first used by \cite{anderson1996method} and \cite{hamill2001interpretation} for ensemble forecasts and then by \cite{cook2006validation} for validating Bayesian inference.
\begin{theorem}
    Let $f \colon \Theta \to \R$ be a measurable mapping and $(\x,\bt) \sim p(\x,\bt)$. Then the rank statistic $r \colon \Theta \to [0,1]$ given by
    \begin{equation*}
        r(\bt) = \ev_{p(\vt \vert \x)}\left[\mathds{1}{\left(f(\bt) < f(\vt)\right)}\right]\label{eq:rank_check_eq}
    \end{equation*}
follows the uniform distribution $\mathcal{U}(0,1)$.
\end{theorem}
If the approximation is accurate, replacing $p(\vt \vert \x)$ by $\hat{p}(\vt \vert \x)$ in Theorem \eqref{eq:rank_check_eq} still gives a uniform distribution on $[0,1]$. As with the coverage check, we need to estimate the ranks in a simulation study. We generate samples $(\x_1,\bt_1),\ldots,(\x_N,\bt_N) \stackrel{\text{iid}}{\sim} p(\x,\vt)$ and for each $\left(\x_i,\bt_i \right)$, we apply an algorithm to generate samples $\vt_1,\ldots,\vt_L \sim \hat{p}{(\vt \vert \x_i)}$ and approximate the rank $r(\bt_i)$ by 
\begin{equation*}
\frac{1}{L}\sum_{l=1}^L \mathds{1}\left[f(\bt) < f( \vt_i) \right].
\end{equation*}
In fact, we recognize the coverage check from Definition \ref{eq:coverage_def} as a particular case of the rank check for $f(\bt) =\hat{p}(\bt \mid \x)$. The per-parameter coverage checks from Section~\ref{subsection:posterior_coverage_and_novel_checks} can also be interpreted as rank checks for $f(\bt) = \hat{p}(\theta^i \mid \bt^{1:i-1}, \x)$, and the Wasserstein metric displayed in Tables \ref{table:calibration_metrics} and \ref{table:calibration_metrics_tre_beta_iso} can be interpreted as the Wasserstein distance between the distribution of posterior ranks and the theoretical uniform distribution.

The rank approximation is exact in the limit $L \to \infty$. Nevertheless, in this particular case, the finite-sample regime can be very different from the asymptotic one. If $\vt_1,\ldots,\vt_L$ were independent, Glivenko-Cantelli would ensure uniform convergence of the empirical cdf; this result does not generally hold for dependent samples \citep{Adams2010UniformCO}. \cite{talts2018validating} show empirically on a linear regression example that MCMC autocorrelation is enough to create deviations of the rank statistics from the uniform distribution and propose 
thinning the MCMC chain by $\floor{L / L_{\text{eff}}}$ for more reliable results, where $L_{\text{eff}}$ is the effective sample size. This further motivates the use of Chebyshev polynomials for posterior sampling, which produces independent samples. Finally, the rank check can be satisfied by poor approximations too, see \cite{zhao2021diagnostics}.

\newpage
\section{Extended simulation study}
\label{SM:extended_sim_study}
\subsection{Point estimators}
\label{sm:point_estimators}
\begin{figure}[t]
   \centering
    \includegraphics[width=\textwidth]{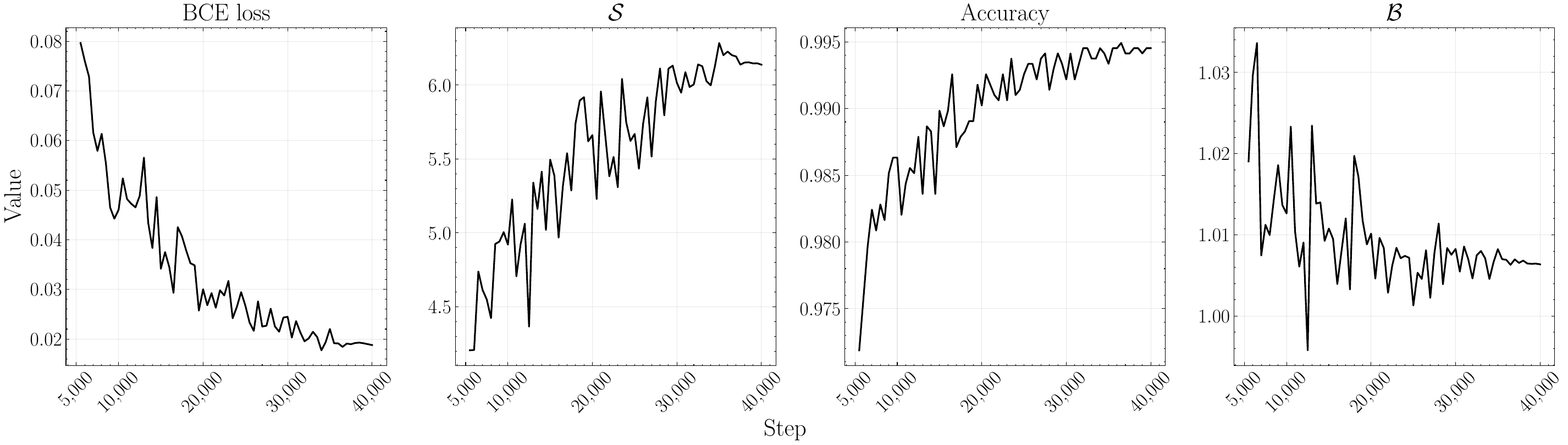}
    \vspace{-1cm}
    \caption{\small{BCE, $\mathcal{S}$, accuracy, and $\mathcal{B}$ metrics (left to right) for the NRE classifier, evaluated on a holdout dataset over the last 35000 training iterations. We train the classifiers with trawl process realizations $\x$ of length $k=1500$.}}
\label{fig:nre_tracking_metrics_during_training_nre}
\end{figure}
We present further details on the NRE , GMM and NBE point estimation methodologies previously used in Section~\ref{subsection:point_estimators}.

\textbf{NRE:} To begin with, we train the NRE in the same way as the TRE, optimizing for the BCE loss and computing gradients from data simulated on-the-fly. We report the training metrics in Figure~\ref{fig:nre_tracking_metrics_during_training_nre}. All metrics stabilize, suggesting numerical convergence.

\textbf{GMM:} Next we discuss GMM estimation for a trawl process realization $\x$. This procedure is theoretically motivated by the fact that the trawl processes is stationary and ergodic, hence moment-based estimation is consistent \citep{barndorff2014integer}. Let $\bt_{\text{a}}$ and $ \bt_{\text{m}}$ be the ACF and marginal parameters, respectively. We infer these parameters separately: $\bt_{\text{a}}$ by matching the theoretical and empirical ACFs, and $\bt_{\text{m}}$ by matching the theoretical and empirical moments of $X$. Formally, define the vectors of moment conditions
\begin{equation*}
    g_{\text{a}}(\bt) = 
    \begin{bmatrix}
        \tilde{\rho}(1) - \rho(1; \bt_\text{a}) \\
        \vdots \\
        \tilde{\rho}(K) - \rho(K; \bt_\text{a}) 
    \end{bmatrix} \quad \text{ and} \quad
        g_{\textrm{m}}(\bt) = 
    \begin{bmatrix}
        \tilde{m}_1(\x) - m_1(\bt_\text{m}) \\
        \vdots \\
        \tilde{m}_J(\x) - m_J(\bt_\text{m})
  \end{bmatrix},
\end{equation*}
where $\tilde{\rho}(k)$ denotes the empirical autocorrelation at lag $k$, $ \rho(k; \bt_\text{a})$ the theoretical autocorrelation at lag $k$, $\tilde{m}_j(\x)$ the $j^\text{th}$ empirical moment, and $m_j(\bt_{\text{m}})$ the $j^\text{th}$ theoretical moment of $X$. The number of lags $K$ and moments $J$ remains to be chosen by the user. The weighted GMM estimators are then given by 
\begin{align*}
    \hat{\bt}_{\text{a}} = \argmin_{\bt} \; g_{\text{a}}(\bt)^\top W_{\text{a}} \, g_{\text{a}}(\bt), \\
    \hat{\bt}_{\text{m}} = \argmin_{\bt} \; g_{\text{m}}(\bt)^\top W_{\text{m}} \, g_{\text{m}}(\bt),
\end{align*}
where $W_{\text{a}}$ and $W_{\text{m}}$ are symmetric positive semidefinite weight matrix. Sometimes, these matrices are chosen to be the identity, for simplicity, resulting in 
\begin{align*}
    \hat{\bt}_{\text{a}} &= \argmin_{\bt} \; \sum_{k=1}^K \big(\tilde{\rho}(k) - \rho(k; \bt_\text{a})\big)^2, \\
    \hat{\bt}_{\text{m}} &= \argmin_{\bt} \; \sum_{j=1}^J \big(\tilde{m}_j(\x) - m_j(\bt_\text{m})\big)^2.
\end{align*}
The optimal choice of $W$ is the inverse of the asymptotic covariance matrix of the empirical ACF at and moments. However, for finite samples, especially when the autocorrelation decays slower than exponential, the choice of $W$ strongly impacts the stability and performance of the estimator. We use the default implementation from Pyhon's statsmodels library, version 0.14.5, with $K=35$ and $J=4$. A thorough summary of the GMM applied to trawl processes can be found in Section S3 of \cite{cl_integer_trawl}. 
\begin{remark}
    In principle, the ACF and marginal parameters can be inferred jointly, but this increases the number of matrix entries to be estimated. We found this to significantly affect performance and degrade results in finite samples.
\end{remark}

\textbf{NBE:} In this setting, we train two separate neural networks. Each network takes as input a realization of the trawl process $\x$ and outputs an estimator of either $\hat{\bt}_{\text{a}}$ or $\hat{\bt}_{\text{m}}$. The main challenge lies in the choice of an appropriate loss function. A natural first attempt is to minimize the MAE or MSE between the true and inferred parameters. However, in practice this approach fails for the ACF parameters: the network output collapses to a constant. We attribute this failure to the  non-identifiability of the ACF functions, in the sense that very different values of $\bt_{\text{a}}$ can produce nearly indistinguishable ACFs. To remedy this issue, for the ACF network, we select the $L^2$ distance between the true and inferred ACFs as loss function $
    \bt_{\text{a}} \to \sqrt{\sum_{k=1}^K \left(\rho(k) - \rho(k; \bt_{\text{a}})\right)^2}$, where $K$ is again to be chosen. For the marginal distribution network, we experiment with several possible loss functions:
    \begin{itemize}
\item MSE between the inferred and true marginal parameters $\bt_{\text{m}}$,
\item Kullback–Leibler (KL) divergence from the true to the inferred marginal distribution,
\item reversed KL divergence (rKL), i.e. the KL divergence from the inferred to the true distribution, as in variational inference,
\item symmetrized KL divergence (sym), defined as the mean of KL and rKL.
\end{itemize}
\begin{table}[t]
\caption{\small{We compare estimation errors across series lengths $k=1000$, $1500$, and $2000$ for GMM, NRE, TRE, and NBE. For the ACF parameters, only a single estimator is considered, and we display the mean $L^1$ and $L^2$ distances between the true and inferred ACFs. For the marginal parameters $\mu, \sigma$, $\beta$, we report results for NBE-MSE, NBE-KL, NBE-rKL, and NBE-sym, corresponding to the specific loss function used during training. In this case, we display the MAE and RMSE between the true and inferred parameters, as well as KL and reverse KL divergences between the true and inferred distributions. Overall, TRE outperforms both NRE and GMM, and performs comparably to the different NBE variants. Note that NBE (MSE) from this table is referred to as NBE in Table~\ref{table:mle}. In Table~\ref{table:mle} we only display results for one NBE only, hence there is no possibility of confusion.}}
\scalebox{0.75}{
\begin{tabular}{l@{\hspace{8mm}}l|rr|rr|rr|rr|r|r}
\toprule
 &  & \multicolumn{2}{c|}{ACF} & \multicolumn{2}{c|}{$\mu$} & \multicolumn{2}{c|}{$\sigma$} & \multicolumn{2}{c|}{$\beta$} & mean KL & mean rKL \\
 &  & mean $L^1$ & mean $L^2$ & MAE & RMSE & MAE & RMSE & MAE & RMSE &  &  \\
\midrule
\multirow{7}{*}{1000} & GMM & 3.473 & 0.615 & 0.224 & 0.335 & 0.248 & 0.323 & 1.390 & 1.963 & 0.444 & 0.426 \\
 & NRE & 1.518 & 0.269 & 0.110 & 0.150 & 0.112 & 0.147 & 0.760 & 1.058 & 0.036 & 0.035 \\
 & TRE & 1.266 & 0.224 & 0.098 & 0.134 & 0.090 & 0.119 & 0.631 & 0.860 & 0.026 & 0.026 \\
 & NBE (MSE) & \multirow{4}{*}{1.218} & \multirow{4}{*}{0.215} & 0.101 & 0.135 & 0.089 & 0.115 & 0.529 & 0.695 & 0.040 & 0.040 \\
 & NBE (KL) &  &  & 0.097 & 0.131 & 0.090 & 0.117 & 0.725 & 0.955 & 0.027 & 0.029 \\
 & NBE (rKL) &  &  & 0.098 & 0.130 & 0.091 & 0.117 & 0.734 & 0.954 & 0.031 & 0.032 \\
 & NBE (sym) &  &  & 0.112 & 0.149 & 0.093 & 0.118 & 0.728 & 0.957 & 0.037 & 0.039 \\
\cline{1-12}
\multirow{7}{*}{1500} & GMM & 3.084 & 0.546 & 0.196 & 0.300 & 0.226 & 0.299 & 1.285 & 1.834 & 0.374 & 0.355 \\
 & NRE & 1.308 & 0.232 & 0.094 & 0.125 & 0.098 & 0.127 & 0.686 & 0.964 & 0.025 & 0.025 \\
 & TRE & 1.071 & 0.190 & 0.082 & 0.112 & 0.078 & 0.101 & 0.554 & 0.764 & 0.017 & 0.017 \\
 & NBE (MSE) & \multirow{4}{*}{1.021} & \multirow{4}{*}{0.180} & 0.087 & 0.114 & 0.077 & 0.098 & 0.458 & 0.604 & 0.029 & 0.029 \\
 & NBE (KL) &  &  & 0.082 & 0.111 & 0.077 & 0.100 & 0.649 & 0.861 & 0.019 & 0.019 \\
 & NBE (rKL) &  &  & 0.081 & 0.109 & 0.073 & 0.094 & 0.585 & 0.779 & 0.017 & 0.017 \\
 & NBE (sym) &  &  & 0.082 & 0.110 & 0.076 & 0.098 & 0.591 & 0.780 & 0.018 & 0.018 \\
\cline{1-12}
\multirow{7}{*}{2000} & GMM & 2.848 & 0.504 & 0.178 & 0.281 & 0.204 & 0.274 & 1.232 & 1.794 & 0.336 & 0.302 \\
 & NRE & 1.192 & 0.211 & 0.084 & 0.113 & 0.090 & 0.117 & 0.636 & 0.911 & 0.021 & 0.020 \\
 & TRE & 0.945 & 0.167 & 0.074 & 0.100 & 0.070 & 0.091 & 0.502 & 0.698 & 0.014 & 0.014 \\
 & NBE (MSE) & \multirow{4}{*}{0.909} & \multirow{4}{*}{0.161} & 0.079 & 0.103 & 0.072 & 0.091 & 0.416 & 0.552 & 0.025 & 0.025 \\
 & NBE (KL) &  &  & 0.074 & 0.100 & 0.072 & 0.092 & 0.603 & 0.807 & 0.015 & 0.015 \\
 & NBE (rKL) &  &  & 0.074 & 0.100 & 0.070 & 0.090 & 0.554 & 0.722 & 0.015 & 0.015 \\
 & NBE (sym) &  &  & 0.078 & 0.105 & 0.072 & 0.092 & 0.551 & 0.717 & 0.017 & 0.017 \\
\cline{1-12}
\bottomrule
\end{tabular}
}
\label{table:mle_with_nbe}
\end{table}

Wheareas back-propagating through the MSE loss is straightforward, the loses containing the KL divergence require special attention. Specifically, there is no closed form expression for the KL divergence between two NIG distributions, and we have to approximate these quantities from samples during training. The reversed KL divergence introduces an additional complication: we must compute gradients of samples from the NIG distribution with respect to the parameters of the NIG distribution we sample from. To enable this, we employ the pathwise gradient (reparameterization trick) as described in \cite{reparam_tutorial}.

We summarize results for GMM, NRE, TRE and NBE in Table~\ref{table:mle_with_nbe}. In the NBE setup, the ACF network is trained with a single loss function, and therefore we report its results once. For marginal inference, we display results under the four losses from above: MSE, KL, rKL and sym. Note that their estimation errors differ slightly depending on the objective function used for training. We use $K=35$ lags to approximate the $L^2$ ACF distance for NBE and GMM both during inference and as evaluation metric. Finally, note that TRE achieves performance comparable to that of NBE, despite this obvious disadvantage. Whereas the NBE and GMM point estimators are trained with the same metric as the one displayed in the table, the point estimates from TRE (and NRE) maximize the likelihood function, i.e., a different criterion. 

\subsection{Beta calibration versus isotonic regression for TRE}\label{SM:extended_beta_vs_iso}
\begin{figure}[t]
    \centering
    \includegraphics[]{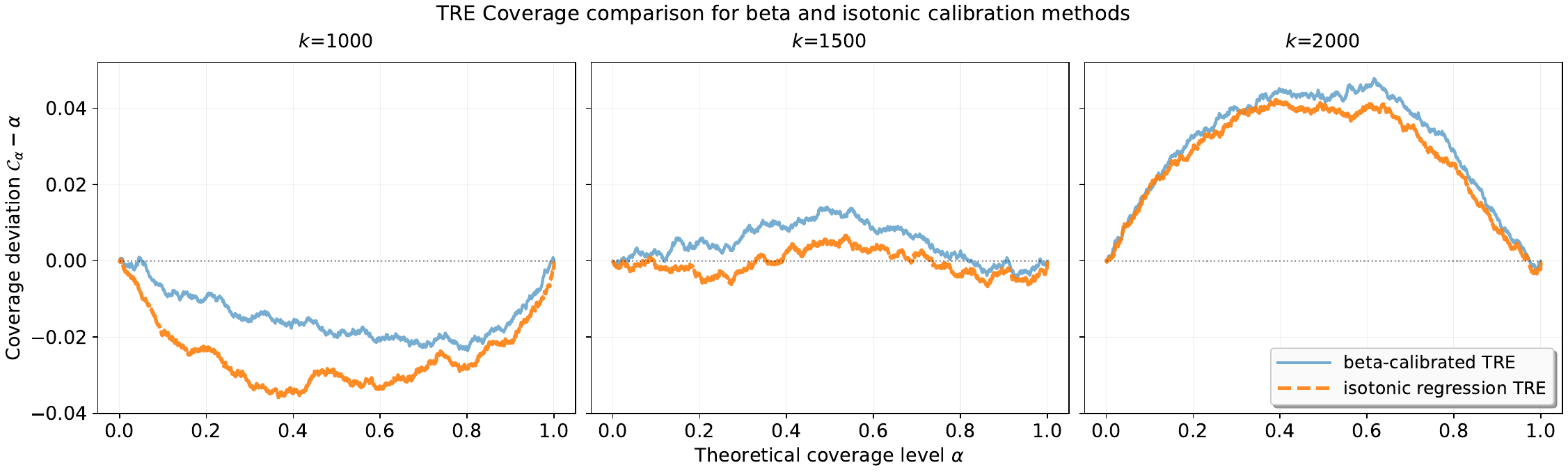}
     \caption{\small{Comparison of the coverage deviation $\mathcal{C}_{\alpha} - \alpha$ for the TRE under beta calibration and isotonic regression. Positive deviations indicate underconfidence, while negative values reflect overconfidence. Isotonic regression reduces coverage deviations relative to beta calibration for $k=1500$ and $k=2000$, but yields worse coverage deviations for $k=1000$.}}
    \label{fig:coverage_comparison_tre_beta_iso}
\end{figure}
We have shown in Section~\ref{section:calibration_coverage_posterior_checks} that beta-calibration improves the quality of the trained TRE and enables amortization over the time-series length $k$. However, we have also observed one case in which beta-calibration degrades performance, the NRE at $k=1000$; see Figure~\ref{fig:coverage_comparison} and Table~\ref{table:calibration_metrics}. We attribute this to the nearly degenerate outputs of the NRE, which are often close to $0$ or $1$ and therefore difficult to calibrate. Fortunately, for the TRE, this is not a big problem: by breaking the initial classification task into multiple smaller tasks, we avoid such degeneracy.

It is worth asking if isotonic regression, which is a non-parametric calibration method, offers improvements over beta-calibration. This would be the case if the family of beta-calibration maps would be too restrictive for the task at hand. As shown in Figure~\ref{fig:coverage_comparison_tre_beta_iso}, isotonic regression improves coverage for $k=1500$ and $k=2000$, but degrades it for $k=1000$. Comparing the metrics in Table~\ref{table:calibration_metrics_tre_beta_iso}, isotonic regression performs better at $k=1500$, while the results are mixed for $k=1000$ and $k=2000$. At $k=1000$, the BCE and $\mathcal{S}$ metrics improve, but $W_1$ worsens; at $k=2000$, $W_1$ improves slightly, while $\mathcal{S}$ degrades slightly. We do not include additional coverage deviation results for the component NREs within the TRE, as the differences are small enough to fall within Monte Carlo simulation error.
\begin{table}[t!]
\caption{\small{Comparison between TRE and component NREs (ACF, $\beta$, $\mu$, $\sigma$) calibrated with the beta-calibration versus isotonic regression. We display the following metrics: BCE, $\mathcal{S}$, $\mathcal{B}$, Wasserstein distance $W_1$, and Expected Calibration Error (ECE). Values are shown across sequence lengths $k = 1000$, $1500$ and $2000$. Isotonic regression performs better for $k=1500$, but we do not have a clear winner for $k=1000$ and $2000$.}}
\centering
\scalebox{0.85}[0.875]{%
\begin{tabular}{l@{\hspace{8mm}}l!{\vrule width 0.8pt}rr!{\vrule width 0.3pt}rr!{\vrule width 1pt}rr!{\vrule width 0.3pt}rr!{\vrule width 0.3pt}rr!{\vrule width 0.3pt}rr}
\toprule
 &  & \multicolumn{2}{c}{TRE} & \multicolumn{2}{c}{ACF} & \multicolumn{2}{c}{$\beta$} & \multicolumn{2}{c}{$\mu$} & \multicolumn{2}{c}{$\sigma$} \\
 &  & beta & iso & beta & iso & beta & iso & beta & iso & beta & iso \\
\midrule
 & BCE & 0.025 & 0.024 & 0.430 & 0.430 & 0.223 & 0.223 & 0.299 & 0.299 & 0.317 & 0.317 \\
 & $\mathcal{S}$ & 6.177 & 6.247 & 0.995 & 1.015 & 2.064 & 2.083 & 1.615 & 1.631 & 1.504 & 1.517 \\
\multirow{-1}{*}{1000} & $\mathcal{B}$ & 0.999 & 0.998 & 1.000 & 1.000 & 0.999 & 0.999 & 1.000 & 1.000 & 0.999 & 1.000 \\
 & $W_1$ & 0.015 & 0.025 & 0.003 & 0.005 & 0.006 & 0.006 & 0.003 & 0.012 & 0.006 & 0.006 \\
 & ECE & --- & --- & 0.009 & 0.002 & 0.006 & 0.001 & 0.007 & 0.001 & 0.006 & 0.001 \\
\midrule
 & BCE & 0.015 & 0.015 & 0.388 & 0.388 & 0.199 & 0.199 & 0.268 & 0.268 & 0.285 & 0.285 \\
 & $\mathcal{S}$ & 6.949 & 6.981 & 1.186 & 1.198 & 2.267 & 2.273 & 1.805 & 1.814 & 1.690 & 1.696 \\
\multirow{-1}{*}{1500} & $\mathcal{B}$ & 1.000 & 0.999 & 1.000 & 1.000 & 1.001 & 1.001 & 1.001 & 1.001 & 1.002 & 1.002 \\
 & $W_1$ & 0.006 & 0.003 & 0.003 & 0.003 & 0.013 & 0.008 & 0.006 & 0.003 & 0.015 & 0.011 \\
 & ECE & --- & --- & 0.002 & 0.002 & 0.002 & 0.001 & 0.002 & 0.001 & 0.004 & 0.002 \\
\midrule
 & BCE & 0.010 & 0.010 & 0.364 & 0.364 & 0.186 & 0.186 & 0.250 & 0.250 & 0.264 & 0.264 \\
 & $\mathcal{S}$ & 7.442 & 7.430 & 1.309 & 1.308 & 2.389 & 2.380 & 1.926 & 1.926 & 1.818 & 1.816 \\
\multirow{-1}{*}{2000} & $\mathcal{B}$ & 1.000 & 1.000 & 1.001 & 1.001 & 1.001 & 1.001 & 1.001 & 1.001 & 1.001 & 1.001 \\
 & $W_1$ & 0.030 & 0.027 & 0.013 & 0.011 & 0.033 & 0.032 & 0.010 & 0.009 & 0.022 & 0.022 \\
 & ECE & --- & --- & 0.004 & 0.002 & 0.003 & 0.001 & 0.003 & 0.001 & 0.003 & 0.001 \\
\bottomrule
\end{tabular}
}
\label{table:calibration_metrics_tre_beta_iso}
\end{table}

\section{Model architectures}
The Python JAX implementation and configuration files are available at 
\url{https://github.com/danleonte/Simulation-based-inference-via-telescoping-ratio-estimation-for-trawl-processes-paper}.
Table~\ref{table:architecture} provides the architectural details, which follow the design shown in Figure~\ref{fig:individual_nre_within_tre}. For optimization, we employ the Adam optimizer with cosine weight decay using JAX Optax 0.2.2, with the hyperparameter $\alpha$ specified in the table below.
\label{section:training_details}
\begin{table}[!b]
\caption{\small{Architectures and training hyper-parameters}}
\centering
\resizebox{\textwidth}{!}{
\begin{tabular}{lccccc}
\toprule
& NRE &  ACF & $\mu$ & $\sigma$ & $\beta$ \\
\midrule
\textit{LSTM encoder hidden size} & 128  & 128 &  128 & 128 & 128 \\
\textit{LSTM layers} & 2  & 2 & 2 & 2 & 2 \\
\textit{Head network} & MLP & MLP & MLP & MLP & MLP \\
\textit{Head layers neurons} & [128,48,32,15,8,4,2] &  [64,32,16,8,4,2] & [64,32,16,8,4] & [128,48,32,15,8,4,2] & [128,48,32,15,8,4,2]  \\
\textit{Learning rate} & $5 \cdot 10^{-4}$ & $5 \cdot 10^{-4}$  & $5 \cdot 10^{-4}$ & $5 \cdot 10^{-4}$ & $5 \cdot 10^{-4}$ \\
\textit{Alpha}&  $2.5 \cdot 10^{-3}$ & $5 \cdot 10^{-3}$  & $5 \cdot 10^{-3}$ & $5 \cdot 10^{-3}$ & $5 \cdot 10^{-3}$  \\
\textit{Dropout rate} & $5 \cdot 10^{-2}$  & $5 \cdot 10^{-2}$ & $5 \cdot 10^{-2}$ & $5 \cdot 10^{-2}$  & $5 \cdot 10^{-2}$ \\
\textit{Iterations} & 40000  & 40000 & 40000 & 40000 & 40000 \\
\textit{Batch size} & 64  & 64 & 64 & 64 & 64 \\
\bottomrule
\end{tabular}
}
\label{table:architecture}
\end{table}

\newpage
\clearpage

\end{document}